%% file: main.tex
\documentclass[acmtog]{acmart}
\acmSubmissionID{392}

\usepackage{booktabs} % For formal tables
\usepackage{rotating}
\usepackage{wrapfig,graphicx}
\usepackage[caption=false]{subfig}
\usepackage{soul}
\usepackage{svg}

\acmPrice{15.00}

\setcopyright{acmlicensed}\acmJournal{TOG}
\acmYear{2021}\acmVolume{40}\acmNumber{6}\acmArticle{1}\acmMonth{12} \acmDOI{10.1145/3478513.3480562}

\setcitestyle{square}

\input{macros.tex}

\begin{document}

% Title. 
% If your title is long, consider \title[short title]{full title} - "short title" will be used for running heads.
\title{AutoMate: A Dataset and Learning Approach for Automatic Mating of CAD Assemblies}

% Authors.
\author{Benjamin Jones}
\affiliation{%
  \institution{University of Washington}
  \country{USA}}

\author{Dalton Hildreth}
\affiliation{%
  \institution{University of Washington}
  \country{USA}}

\author{Duowen Chen}
\affiliation{%
  \institution{Columbia University}
  \country{USA}}

\author{Ilya Baran}
\affiliation{%
  \institution{PTC Inc.}
  \country{USA}}
  
\author{Vladimir G. Kim}
\affiliation{%
  \institution{Abobe Inc.}
  \country{USA}}

\author{Adriana Schulz}
\affiliation{%
  \institution{University of Washington}
  \country{USA}}

\renewcommand{\shortauthors}{Jones et al.}

\input{sec/0_abstract}

%CCS


%keywords

\input{fig/teaser.tex}
\maketitle
%\ilya{Keywords need updating}
\input{sec/1_introduction.tex}
\input{sec/2_background.tex}
\input{sec/3_overview.tex}
\input{sec/4_dataset.tex}
\input{sec/5_methods.tex}
\input{sec/6_results.tex}
\input{sec/7_discussion.tex}

\bibliographystyle{ACM-Reference-Format}
\bibliography{main,manual}
\input{sec/8_appendix}

\end{document}

%% file: macros.tex
\theoremstyle{definition}

\newcommand{\rel}{{\,:\,}}

\DeclareMathOperator\MLP{MLP}
\DeclareMathOperator\concat{cat}

%% file: sec/0_abstract.tex
\begin{abstract}

Assembly modeling is a core task of computer aided design (CAD), comprising around one third of the work in a CAD workflow. Optimizing this process therefore represents a huge opportunity in the design of a CAD system, but current research of assembly based modeling is not directly applicable to modern CAD systems because it eschews the dominant data structure of modern CAD: parametric boundary representations (BREPs). CAD assembly modeling defines assemblies as a system of pairwise constraints, called \emph{mates}, between parts, which are defined relative to BREP topology rather than in world coordinates common to existing work.  We propose SB-GCN, a representation learning scheme on BREPs that retains the topological structure of parts, and use these learned representations to predict CAD type mates. To train our system, we compiled the first large scale dataset of BREP CAD assemblies, which we are releasing along with benchmark mate prediction tasks. Finally, we demonstrate the compatibility of our model with an existing commercial CAD system by building a tool that assists users in mate creation by suggesting mate completions, with 72.2\% accuracy.

\end{abstract}

%% file: fig/teaser.tex
% A "teaser" figure, centered below the title and authors and above the body of the work.
\begin{teaserfigure}
  \centering
  \includegraphics[width=\textwidth]{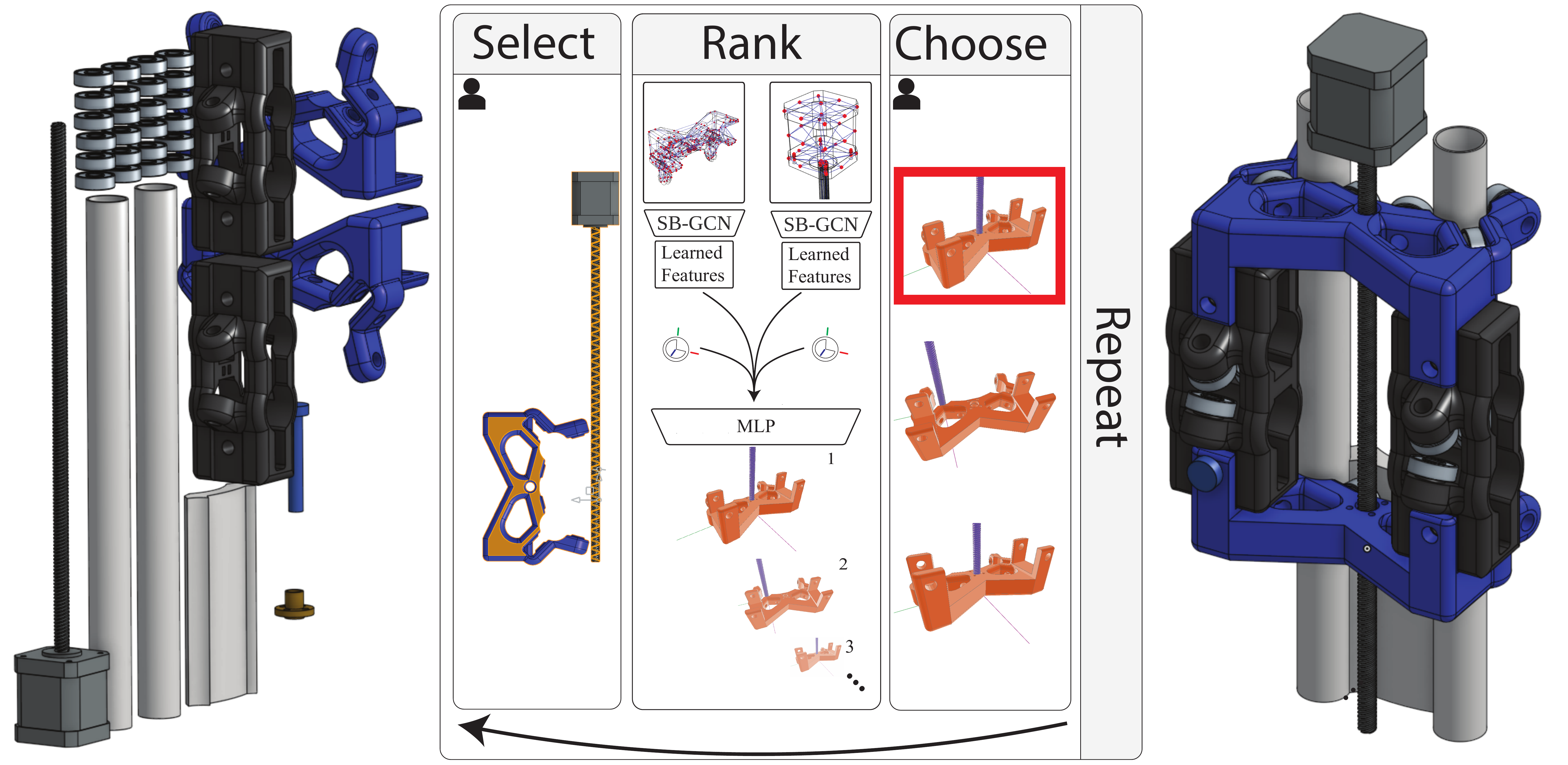}
  \caption{Our tool assists a designer in assembling a collection of CAD parts (left) into a functional assembly (right). First user selects a pair of parts to mate, and indicates roughly where the parts should be mated by where they click on each part. SB-GCN then creates embeddings for each topological entity of the selected parts' BREPs, which are used as input to a classifier that scores potential mates defined in reference to these topological entities. Our system presents a list of suggestions that the user can pick from. This cycle is repeated until a working mechanical assembly is realized. }
\end{teaserfigure}

%% file: sec/1_introduction.tex
\section{Introduction}

\begin{figure*}[t]
    \centering
    \includegraphics[width = \textwidth]{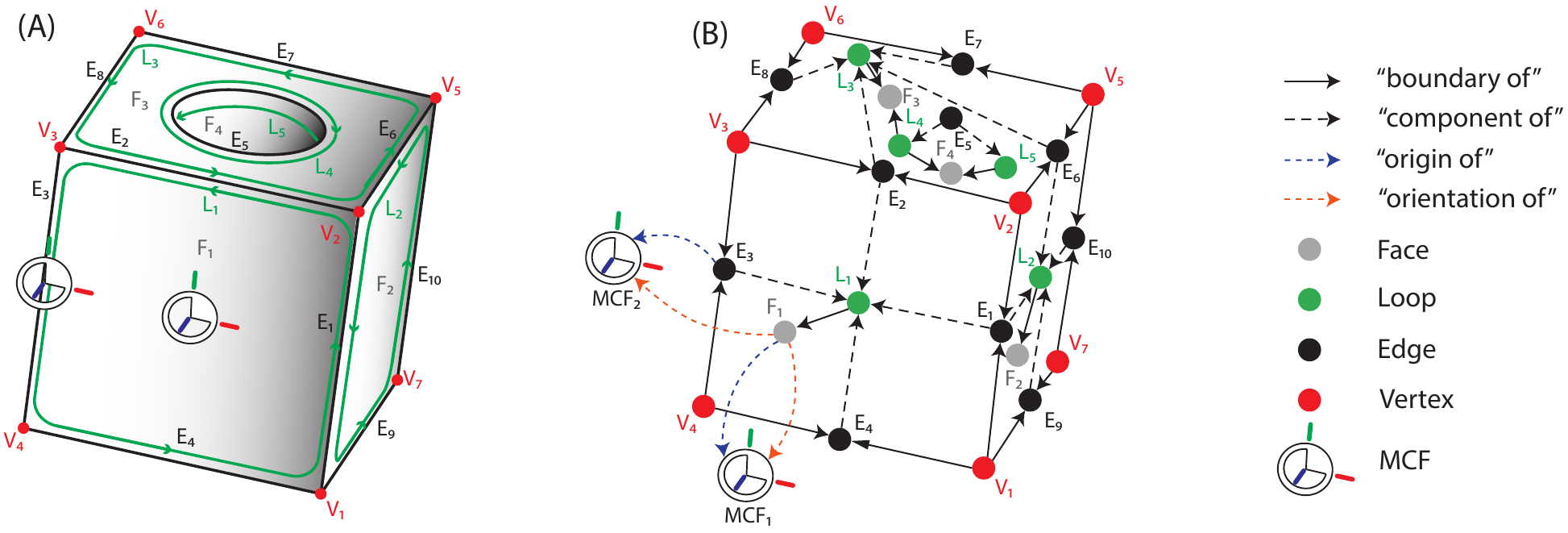}
    \caption{Our data model. Left: A simple part modeled as a BREP, with the topological entities (faces, loops, edges, and vertices) labeled. Also shown are two mating coordinate frames, located at the centroid of $F_1$ and the center of $E_3$. Right: Our graph representation of a BREP; a subset of the graph for the part on the left, limited to visible entities, is shown. The topological entities (graph nodes) are connected via boundary or component relations (graph edges). This graph has consistent structure; graph edges only go vertex-to-edge, edge-to-loop, or loop-to-face. Mating Coordinate Frames are defined by reference to topological entities for their location and orientation.}
    \label{fig:datamodel}
\end{figure*}

Manufactured objects are designed
in parametric computer-aided design (CAD) systems as individual mechanical
parts that are assembled into functional objects. In CAD, creating an assembly
from parts is a time-consuming manual process of \emph{mating} the parts, which requires both careful positioning of parts in reference to one another, as well as specification of how these parts can move relative to one another. In practice, users spend roughly one third of their time in modern CAD tools in assembly work~\cite{onshapecommunication}.
In this work we develop novel machine learning techniques for predicting how to mate parts. We further integrate these predictive models into an existing commercial CAD product~\cite{onshape}, making it easier to create mates.

While many prior techniques use machine learning in assembly-based 3D modeling tools~\cite{sung_complementme_2017,zhu_scores_2018,yin_coalesce_2020,huang_generative_2020,lin_learning_2018,wu_sagnet_2019,mo_structurenet_2019,mo_structedit_2020}, these techniques are not suited for a CAD workflow. Established commercial CAD systems are built around boundary representations (BREPs, Figure~\ref{fig:datamodel}), a data structure that represents the structure of a shape as \emph{topological entities} of different spatial dimensions (2D faces, 1D edges and loops of edges, and 0D vertices), and explicitly captures the \emph{analytic} geometry of each as a parametric equation, as well as the relations between topological entities. BREP based CAD systems model assemblies as a collection of pairwise constraints between parts called \emph{mates}, where the constraints are defined relative to the topological entities. Using an analytic shape representation is crucial for the high modeling accuracy required for fabrication and integration with computed aided manufacturing (CAM). Modeling as a system of constraints enables capturing complex part interactions of mechanical motion, and allows the model to adapt globally to local changes. Defining these constraints from the topological entities allows mates to have analytic precision, as well as adapt to some parametric changes in geometry. In contrast to these CAD native assemblies, many existing assembly modeling tools use segmented meshes or point clouds to represent geometry, and predict world-coordinate, often static positions to construct assemblies, making them ill-suited for integration into the CAD design process.

While some work uses pairwise constraints~\cite{huang_generative_2020, lin_learning_2018, mo_structurenet_2019}, to the best of our knowledge no existing assembly modeling systems create topologically defined mates, nor do any operate on BREP CAD data. Two key challenges impede the development of such a system; the lack of BREP learning models suitable for mating, and a lack of data for training such a system. Concurrent work on learned models for BREPS~\cite{willis_fusion_2020,lambourne_brepnet_2021} operate on a simplified \emph{homogeneous} BREP structure and only learn representations for faces. BREPs can be mated relative to any class of topological entity, so the mating problem requires a \emph{heterogeneous} BREP representation. Compiled BREP datasets~\cite{koch_abc_2019,willis_fusion_2020,lambourne_brepnet_2021} are focused on BREP modeling and do not contain an mate data, and assembly modeling datasets~\cite{mo_partnet_2019} do not have BREP data or mates, so no suitable dataset exists for training such a model.

\begin{figure}[t]
    \centering
    \includegraphics[width = \linewidth]{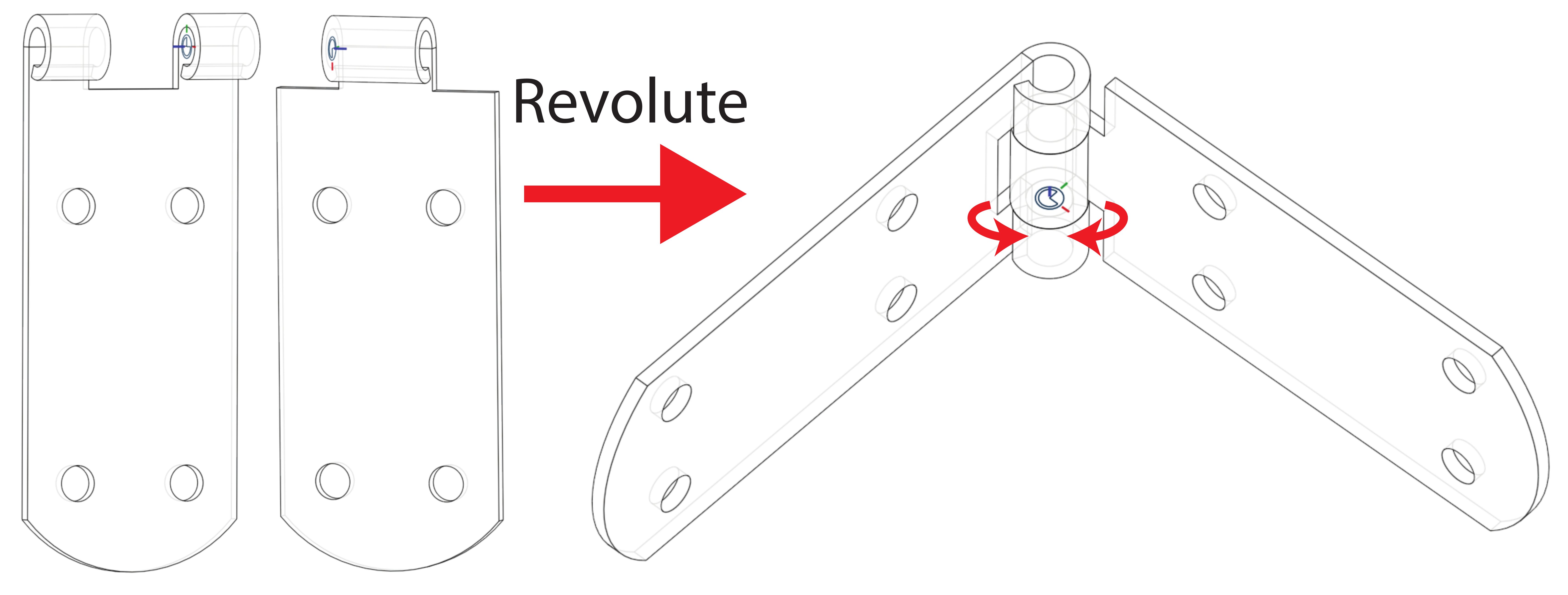}
    \caption{
    In a mate, the references to topological entities specify the location constraints, and the mate type (revolute) determines the degrees of freedom.
    }
    \label{fig:mateexample}
\end{figure}

To overcome these challenges, we offer methodological and data contributions. First, we propose \emph{Structured BREP Graph Convolution Networks} (SB-GCN), a structured graph representation and message passing network for BREPs that is the first heterogeneous learned BREP representation in that it learns embeddings for multiple classes of topological entity. Second, we scraped a large collection of raw BREP assemblies, and cleaned it by de-duplicating mates and assemblies and converting to a canonical format, to create the first BREP assembly and mating dataset, which we are making publicly available. We apply SB-GCN to BREP mating by modeling mates as translational and rotational constraints between \emph{mating coordinate frames} (MCFs) -- coordinate frames on BREPs that inherit their origin and alignment by reference to separate topological entities on each BREP (see Figure~\ref{fig:mateexample}).

We demonstrate that our approach is directly applicable to existing commercial CAD system by implementing a mate recommendation system as an extension to the Onshape CAD system. Our mate model maps directly onto 180,102 of the mates in our dataset, which we use to train a ranker for potential mates and a classifier for mate degrees of freedom. Our model suggests a verifiably correct mate location 72.2\% of the time on this data, significantly outperforming traditional discrete geometry approaches (46.3\%). 

%% file: sec/2_background.tex
\section{Related Work}

\paragraph{Assembly-based Modeling} Assembly-based, or compositional, modeling tools create 3d models by aggregating parts from 3d shape collections. Modeling by example~\cite{funkhouser_modeling_2004} opened this field with two main subproblems; how to choose which parts to combine and how to combine them. The first problem is commonly solved by text search~\cite{funkhouser_modeling_2004}, similarity search~\cite{funkhouser_modeling_2004, chaudhuri_data-driven_2010}, or consistent segmentation~\cite{Chaudhuri2011}. In our work, we assume that retrieval is solved and focus solely on the second problem of mating parts.

Traditionally, aligning parts has been performed analytically; using iterative closest points~\cite{funkhouser_modeling_2004, sharf_snappaste_2006}, surface registration~\cite{chaudhuri_data-driven_2010}, or energy optimization \cite{gonzalez_3d_2018}. Newer data-driven methods learn to align shapes. In contrast to our work, most approaches~\cite{ yin_coalesce_2020, huang_generative_2020, sung_complementme_2017,jones_shapeassembly_2020,wu_pq-net_2020,vedaldi_pt2pc_2020,yang_dsm-net_2020} predict or generate global offsets for each part rather than pairwise constraints. Assemblies defined by global positions cannot handle replacement of parts, parametric variation of parts, or modeling degrees of freedom. Those that do model assemblies by pairwise positioning~\cite{zhu_scores_2018,lin_learning_2018, wu_sagnet_2019, mo_structurenet_2019, mo_structedit_2020} predict or generate explicit offsets and rotations, forfeiting the precision and adaptability to parametric changes that our system acquires by defining these relative to topological features. 

Another limitation of these methods is that they work with non-parametric surface representations (meshes, pointclouds, voxel grids) or bounding boxes, and encode the geometry using a geometric learning model such as PointNet~\cite{qi_pointnet_2017}, PointNet++~\cite{qi_pointnetpp_2017}, PCPNet~\cite{guerrero_pcpnet_2018}, or DGCNN~\cite{wang_dynamic_2019}; we refer the reader to Bronstein et. al~\shortcite{bronstein_geometric_2017} for a survey geometric learning methods. Unlike these techniques, our method uses BREPs as input which provides additional CAD context and enables it to interface with CAD systems.

Our work overcomes limitations of existing assembly modeling methods in the CAD context. Existing methods are trained over categories of shapes, limiting their usefulness for general design problems. Since we learn local representations on the underlying part structure, we can avoid dependence on global part context. Additionally, none of these systems model the degrees of freedom of connections. Fab-by-example~\cite{schulz2014design} is the closest existing work to our system in capabilities; its pairwise connections are parametrically defined to adapt to parametric variation, it models the degrees of freedom of connections, and is not limited to a single category. However, it relies on a database of hand-annotated parts and so does not generalize to arbitrary CAD data.%\ben{Argument about local part structure added in response to Vova's request for a rationale as to why we don't need categories, but I think it colud open the door to reviewers wanting us to compare against part stitching methods more than placement methods (these will use local shape energies or surface descriptors to refine alignments and stitch surfaces)}

% Re-write lost \cite{dubrovina_composite_2019}, and \cite{yi_deep_2018}

\paragraph{Learning with CAD Data}
Recently, interest in data-driven approaches on parametric CAD data has increased with the creation of several CAD datasets. Koch et. al released the ABC dataset~\cite{koch_abc_2019}, a collection of CAD models sourced from Onshape, to spur development of learning on CAD data. Sketchgraphs~\cite{seff_sketchgraphs_2020} is a dataset that collects 2D sketch information from Onshape. That work also presents an application in learning the constraint graph underlying these sketches. Recently, \cite{willis_fusion_2020} compiled a dataset of sketch-extrude modeled parts for learning 3D construction sequences, and \cite{cao2020graph} proposed a method of generating CAD datasets with segmentation annotations, and a message passing network for segmentation. Concurrent to our work, \cite{lambourne_brepnet_2021} and \cite{jayaraman2021uv} propose message passing networks to learn BREP face embeddings. Similar to this paper, these works use message passing graph neural networks to model the CAD data, but they only learn to embed one type of topology; sketches in the 2d case and faces in the 3d works. Our mating task requires us to have representations for faces, loops, edges, and vertices, since all 4 classes of topological entity can be used to mate parts. This means our graph representations are inherently heterogeneous, in constrast to the homogeneous graph representations used in prior works.
These datasets and applications are centered around the part modeling part of the CAD workflow. An assembly database, PartNet~\cite{mo_partnet_2019}, does exist, and was used to train several of the assembly modeling systems mentioned earlier, but it is based around mesh geometry, not CAD models. Ours is the first dataset and application that captures assembly information - the relationships between parts. 

\paragraph{Graph Network Representation Learning}
Generic graph-based learning techniques have been applied to geometry, part assemblies, social networks, and other types of data. In this work we leverage these techniques to process CAD data by applying them to a graph structure that takes advantage of the structure of BREPs. We refer the reader to two recent surveys for a more in-depth overview of the field, ~\cite{wu_comprehensive_2020}, ~\cite{zhang_deep_2020} and only reference the most relevant works here. Most modern works follow the Graph Convolutional Approach (GCN) of Kipf and Welling ~\shortcite{kipf_semi-supervised_2017}, which generalizes the Euclidean convolution to graph domains via a two step message passing process: features of neighboring nodes are collected as computed messages, then these messages are aggregated locally using an order-independent aggregation function. Many techniques from CNNs have analogs in GCNs. Li et. al.~\shortcite{li_deepgcns_2019} showed that residual connections are crucial for training deep GCNs; we incorporate these into our network design since GCN depth controls how large of a local neighborhood our learned representation for each topological feature can incorporate. A novel feature of graph domains that is pertinent to CAD data is domain heterogeneity. While basic GCNs handle structural heterogeneity of the domain, there is no natural extension of CNNs or RNNs that captures discrete typings of nodes and edges, which may have heterogeneous labels. \cite{zhang_deep_2018} introduced meta-paths, typed edges connecting all paths with the same node types, and proposed learning a separate message function for each type of meta-path. Learning heterogeneous messages has the disadvantage of greatly expanding model complexity. Our model incorporates both heterogeneous messages and meta-edges, but we leverage the consistent connectivity structure of BREPs to avoid the increase in model complexity usually associated with typed messages by splitting these into a series of homogeneous message passing layers.

%% file: sec/3_overview.tex
\section{Overview}\label{sec:overview}

\begin{figure}[t]
    \centering
    \includegraphics[width=0.5\textwidth]{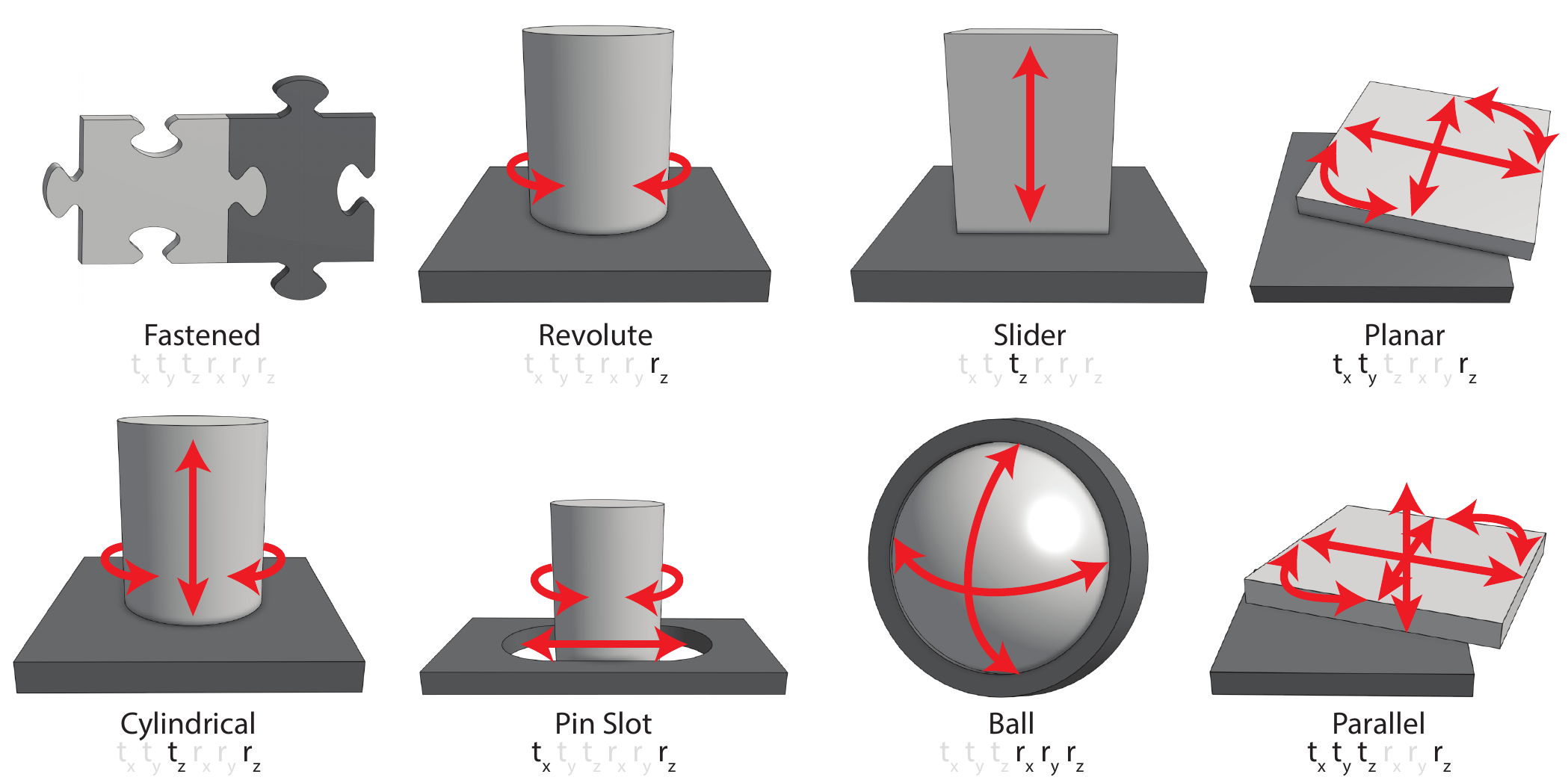}
    \caption{The 8 mate types. Red arrows illustrate the available degrees of freedom for each mate type.}
    \label{fig:matetypes}
\end{figure}

\paragraph{Modeling Mates}
Our predictions are generated by scoring and ranking a collection of potential mates. We model a mate as a pair of \emph{mating coordinate frames} (MCFs), a coordinate frame defined for each mated part in reference to their topological entities (such as faces or edges), and a constraint on their relative position and orientation. For example, the two halves of a hinge in Figure~\ref{fig:mateexample} are mated by MCFs defined in reference to cylindrical faces. The pair of MCFs, one on each part, define the \emph{mate location}, assembling the parts by aligning the frames.  The constraint between MCFs is specified as one of eight \emph{mate types} (Figure~\ref{fig:matetypes}), which specify which axes the frames can be translated or rotated about relative to one another. For example, Figure~\ref{fig:mateexample} shows how a hinge is modeled by a \emph{revolute} mate that disallows translation of the parts, but allow rotation around their common z-axis, letting the hinge swing open and shut.

Each MCF is defined as a tuple of two topological entities from the associated part. One referenced topology defines the origin of the MCF. Some topological entities have more than one location to anchor an MCF to (such as the top, bottom, or middle of a cylinder), so this reference has an associated \emph{origin type} to specify where on the entity to place the origin.\footnote{A complete list of origin types is given in Table~\ref{tab:origintypes}.} The other referenced topological entity determines the frame's orientation by specifying its z-axis, for example, normal to a referenced planar face. The other coordinate axes are derived from the frame each part was modeled in. See Appendix~\ref{ap:datamodel} for specifics. 

This model is simplified from the raw dataset. We discarded a small number of mates that were defined with additional rotational and translational offsets to their MCFs. We also ignore optional per degree-of-freedom range limits, as these are also rare. The raw data we collected also allowed for the MCFs in a mate to be reoriented by one of 8 orthogonal rotations, and for one of the same 8 rotations to be applied to the mate as a whole, giving 512 possible combinations. We canonicalized these down to 8 unique variations for our dataset, and for the models in this paper we limit ourselves to the most common \emph{canonical orientation}.

\paragraph{Interactive Recommendation System}

\begin{figure}
    \centering
    {
    \setlength{\fboxsep}{0pt}
    \fbox{\includegraphics[width = \linewidth]{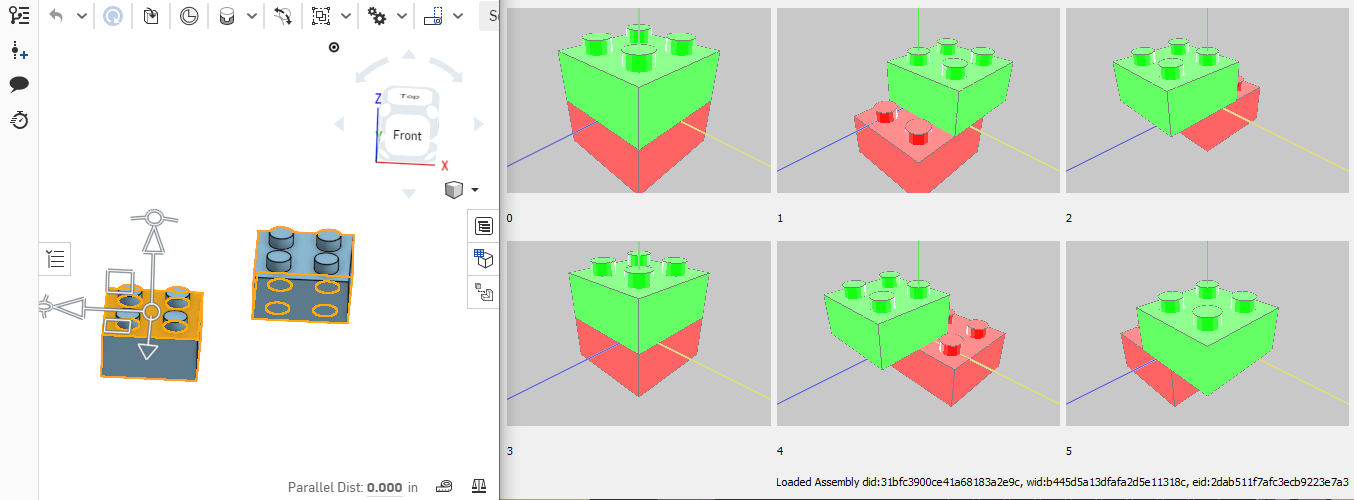}}
    }
    \caption{Our Onshape Extension. Left: Onshape assembly with two parts selected. Right: Top 6 predicted mate locations. The first and fourth suggestions look similar because differently defined mates can resolve to the same location; the first option is centered on middle of the blocks while the fourth is centered on the front peg.}
    \label{fig:ui}
\end{figure}

Our method extends a CAD system to provide autocomplete-like suggestions for mates. Automatically suggesting mates is a balance between automation and control from the user. The problem of deciding how to mate BREPs is inherently ambiguous; it depends on the user's design intent, so there is no ground truth. We decided to structure our tool as a recommendation system. In our system, the user clicks on two parts to select them for mating. We use the face that the user clicked as an indication of design intent; the origin of each MCF will be in the neighborhood of the selected faces. We accomplish this by using the selected face as the orientation reference for each MCF, and restrict our search to MCFs with origin on that face or one of its boundaries. This both captures the user's intent and reduces the cost of the search, and is well supported by our dataset; almost all of the origin references in our dataset are within the boundary of the corresponding orientation topological element. Given this input; two BREPs and face selections on each, our system scores and ranks all pairs of MCFs neighboring the selection, and reports the top 6 to the user. Given a particular MCF pair, our system can also predict the most likely mate types. Figure~\ref{fig:ui} shows our interface for the simple case of two building blocks. The user has selected the top and bottom face of each block to be mated within the system's normal assembly UI. Whenever two faces are selected, our system presents 6 mate location options that the user can choose between with a hotkey, or ignore to continue assembly manually without the autocomplete impeding their workflow. If the user selects a mate suggestion, a \emph{fastened} mate is inserted between the parts, and the user can then select a mate type. Our system also ranks the mate type likelihood conditioned on the chosen mate location.

\paragraph{Predicting Mates}

We score and classify mates using SB-GCN, our graph neural network for BREPs. This message passage network is applied to each input BREP to produce representations for each topological entity that encode the local neighborhood, as well as a global descriptor for each part. Each MCF is encoded by the representations of each referenced entity, the origin type, and the global part descriptor, then we train a classifier to score each MCF (the \emph{mate location task}), or classify it's mate type (the \emph{mate type task}). We train this model over all compatible mates from our dataset. Separating the BREP representation learning from the mate classification model is important because while each CAD system has its own constraint model for mates, most use BREPs as their underlying shape representation and construct their mate by reference to topological entities. By explicitly learning to represent each type of topological entity, our system can be adapted to integrate into other CAD systems.

%% file: sec/4_dataset.tex
\section{Dataset}\label{sec:dataset}

We scraped all publicly available documents from Onshape using their API. We collected 255,213 assemblies with at least one mate, totaling 3,989,164 mates. However, a significant number of these are complete or partial duplicates created by reuse of common hierarchical sub-assemblies, and by copying and modification of assemblies in this public repository.

\begin{figure}
    \centering
    \includegraphics[width=\linewidth]{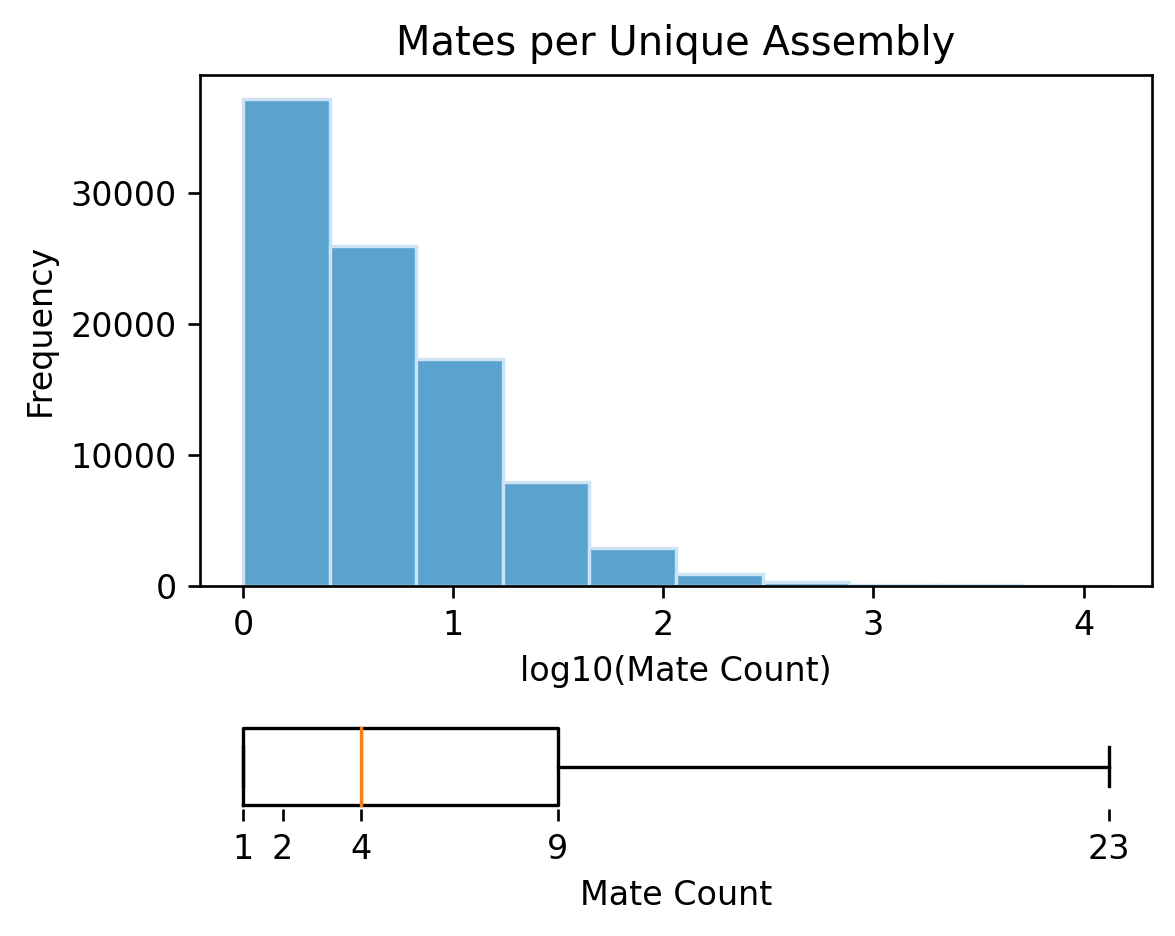}
    \caption{Number of mates per assembly in the dataset. Top: histogram of log assembly size. Bottom: Box-plot of assembly sizes with 10th, 25th 50th, 75th, and 90th percentiles labeled.}
    \label{fig:assemblysizes}
\end{figure}

To clean this dataset, we first expanded and then flattened all sub-assembly references, and filtered down to only top level assemblies. We also removed all incomplete mates, and pairs of parts with multiple mates between them. To identify identical parts and assemblies, we created a BREP fingerprint from the number of each type of topological entity and the moment of inertia tensor and center of mass of the represented solid. We used this to deduplicate re-used parts, and to identify duplicate mates by MCFs, mate types, and mated parts. Assemblies were deduplicated by their mates. The dataset contains 92,529 unique top level assemblies. These unique assemblies range in size from 1 to 13183 mates, with an average of 12, and a 90th percentile of 23. Figure~\ref{fig:assemblysizes} shows the distribution of mates per assembly.

At the mate level, we have 541,635 unique mates. Table~\ref{tab:dsmatetypes} gives the frequency of each mate type among unique mates. Our dataset contains a total of 376,362 unique BREP models used in mates.

\begin{table}[]
    \centering
    \caption{Mate type frequencies.}
    \begin{tabular}{c|c}
         Mate Type & Frequency  \\
         \hline
        FASTENED	&	62.1\%\\
        REVOLUTE	&	12.7\%\\
        PLANAR	&	11.8\%\\
        SLIDER	&	5.3\%\\
        CYLINDRICAL	&	5.1\%\\
        PARALLEL	&	1.8\%\\
        BALL	&	0.6\%\\
        PIN SLOT	&	0.5\%
    \end{tabular}
    \label{tab:dsmatetypes}
\end{table}

%% file: sec/5_methods.tex
\section{Methods}\label{sec:methods}

\paragraph{Graph Representation Learning for BREPs}
BREPs encode both the geometry and the topology of a shape as a collection of topological entities with associated parametric geometry. Faces, edges, and vertices are topological entities associated with parametric geometry functions of surfaces (2D), curves (1D), and points (0D). While the \emph{parameters} (radius, normal, half-angle; see Appendix~\ref{ap:datamodel} for a full list) of these functions are defined for each topological entity, the domain on which to evaluate each function is not. This information is encoded by the \emph{topological} structure of the BREP, defining relationships between topological entities. Lower dimensional entities are related to higher dimensional neighbors by a \emph{boundary of} relationship, and their geometry implicitly defines the domain bounds of these neighbors. A fourth type of topological entity, loops, consist of closed paths of edges. Edges are related to loops by a \emph{component of} relation, and are themselves a boundary of a face. Loops have one parameter to specify if they are an inner or outer boundary of a surface. Figure~\ref{fig:datamodel} (A) shows a simple BREP with faces bounded by loops composed of edges bounded by vertices.

The \emph{boundary of} and \emph{component of} relations define a directed graph in which the topological entities are \emph{nodes}, pictured in Figure~\ref{fig:datamodel} (B). This graph is \emph{heterogeneous} in both its node types and relation types.\footnote{We use \emph{node} and \emph{relation} to refer to \emph{graph} vertices and edges throughout to avoid confusion with  topological entity types of the same names.} Recent and concurrent works~\cite{willis_fusion_2020,lambourne_brepnet_2021} have used a simplified, \emph{homogeneous} version of this graph with only face nodes to build message passing networks for learning BREP representations. These representations are insufficient for learning how to mate BREPs however, because BREP mates can be defined relative to topological entities of any type. Our representation learning must therefore handle the full, heterogeneous BREP graph.

\begin{figure}
    \centering
    \includegraphics{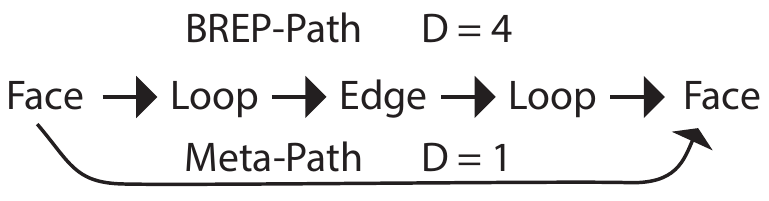}
    \caption{The BREP graph distance between neighboring faces is 4; with meta-paths, this is reduced to 1, shrinking the graph diameter by about 4 times.}
    \label{fig:metapaths}
\end{figure}

Introducing additional nodes and relations presents additional challenges to representation learning. Having multiple types of bounds increases the path length between topological faces; for example, geometrically neighboring face nodes are at graph distance four. We would like our learned embeddings to capture boundary information and local structure, so we want a wide receptive field for our convolutions. In graph convolutional networks, the receptive field is a function of the number of convolutional layers, but as \cite{li_deepgcns_2019} point out, state-of-the-art GCNs are typically only 3-4 layers deep as deeper graph networks are difficult to train. This would barely reach the neighboring face! (See Figure~\ref{fig:metapaths}.)

We adopt two strategies to overcome this. First, we use residual connections in our network, which~\cite{li_deepgcns_2019} showed help stabilize the training of deep GCNs. We also add undirected meta-paths~\cite{zhang_deep_2018} of (face$\leftarrow$loop$\leftarrow$edge$\to$loop$\to$face) relations to connect geometrically adjacent faces and reduce the graph diameter as shown in Figure~\ref{fig:metapaths}. These match the undirected edges of \cite{willis_fusion_2020}.

\begin{figure*}[t]
    \centering
    \includegraphics[width=\textwidth]{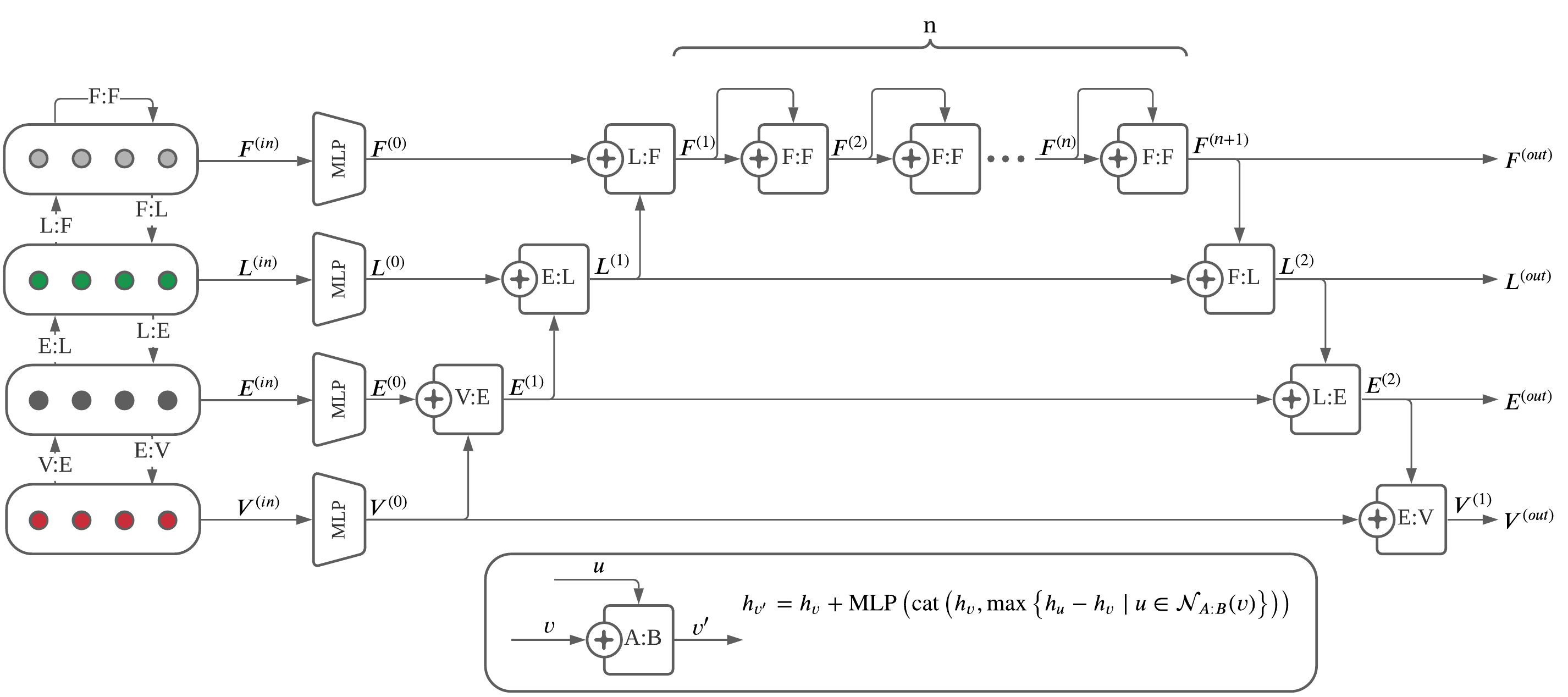}
    \caption{The SB-GCN architecture. BREP graph nodes are structured into 4 tiers by their topological entity type, with boundaries below the type they bound. Boundary information is passed up the tiers from vertices through to faces via residual MRGCN convolutions. Additional undirected MRGCN convolutions between faces quickly widen the network's receptive field, then the aggregated neighborhood information is propagated back down the tiers from faces through to vertices via more convolutions. Separating the convolutions into layers allows each convolutional layer to be homogeneous while still learning separate message passing functions for each type of graph relation.}
    \label{fig:architecture}
\end{figure*}

Learning a different message passing function for each type of relation adds significant time and memory complexity to the model, and so limit the number of layers we can add and the size of our representation. The complexity of a BREP neural network is important to control because BREPs range wildly in complexity, so a large and complex part can easily exhaust available GPU memory. BREPs have a very regular connectivity structure that we use to partition the graph into a \emph{Structured BREP Graph} that we can use to get the advantages of a wide receptive field, heterogeneous graph network for less than the cost of an equivalent homogeneous network.

Since each type of node only has one type of relation, to exactly one other type of node, we can separate our BREP graphs into four \emph{tiers} by node type, as pictured on the left side of Figure~\ref{fig:architecture}. We break our heterogeneous message passing into separate homogeneous message passing layers between adjacent tiers. We first pass messages up through the layers from vertices to faces to give each topological entity its boundary information. Then we perform several layers of message passing over just the face-face meta-paths to widen the receptive field, collecting neighborhood information at each face. Finally, we pass this neighborhood information backwards \emph{down} the tiers from faces to vertices, so in the end, all topological entities have incorporated information about their boundaries and their local neighborhood.

Splitting the heterogeneous convolution into several smaller homogeneous ones saves a significant amount of memory (one sixth the number of parameters) without losing any information on the upward or downward passes. Messages passed from, for example, a loop to a face, can only contain incrementally more information until the information from the vertices at the bottom of the structure reaches the loop anyway, so it is more efficient to incrementally accrue that information in a single upward pass. We call our tiered graph network a \emph{Structured BREP Graph Convolutional Network} (SB-GCN).

\paragraph{SB-GCN}
Formally, we define a Structured BREP Graph $G$ as a collection of node sets, $N = F \cup L \cup E \cup V$, where $F$, $L$, $E$, and $V$ are the face, loop, edge, and vertex nodes respectively. These nodes are then connected by directed, bipartite relation sets $V\rel E$, $E\rel L$, and $L\rel F$ and their transposes $F\rel L$, $L\rel E$, and $E\rel V$. For, say, $V\rel E$ the set contains relations $r_{uv}$ that connects the nodes $u \in V$ to $v \in E$. There is also a set of undirected relations $F\rel F$ which are the meta-path relations between geometrically adjacent faces, computed by vertex-elimination on non-Face nodes in the graph.
SB-GCN takes $G$ and (possibly heterogeneous) input feature vectors on the nodes of $G$ as $N^{(in)} = V^{(in)} \cup E^{(in)} \cup L^{(in)} \cup F^{(in)}$ and produces output feature matrices $V^{(out)}$, $E^{(out)}$, $L^{(out)}$, and $F^{(out)}$ via a series of structured graph convolutions. Figure~\ref{fig:architecture} shows the network architecture.

First a shared MLP transforms all input feature vectors into a common feature space prior to convolution:
\begin{equation}
    N^{(0)} = MLP^{(in)}\left(N^{(in)}\right).
\end{equation}

Next a cascade of residual graph convolutions are computed. We adapt the Max-Relative GCN (MRGCN) that \cite{li_deepgcns_2019} found effective for training deep GCNs with residual connections:
\begin{align}
    \Sigma^{(l)}_{v} = \max\left(\left\{h_u^{(l)}-h_v^{(l)} \mid u \in \mathcal{N}_{\tau(u)\rel\tau(v)}(v)
    \right\}\right) \notag\\
    h_v^{(l+1)} = h_v^{(l)} + \MLP^{(l)}\left(\concat\left(h_v^{(l)}, \Sigma^{(l)}_{v}\right)\right).
\end{align}
This is computed for all $h_v^{(l)} \in N^{(l)}$ where $h_v^{(l)}$ is the feature of the node $v \in N^{(l)}$. $\tau(v)$ is an operator for $v \in N^{(l)}$ which returns which of $F$, $L$, $E$, or $V$ that $v$ is in, that is the topological type of the node $v$, and so $\mathcal{N}_{\tau(u)\rel\tau(v)}(v)$ is the neighborhood of $v$ using the relation set $\tau(u)\rel\tau(v)$.
$\MLP^{(l)}$ is then the $l$th layer's MLP of a linear layer, batch normalization, and ReLU.
The important difference in our convolution is that the adjacency of relations is determined depending on the layer, denoted by the subscript of the neighborhood operator. For the first three layers, relations are in the order of the BREP hierarchy: Vertex to Edge, Edge to Loop, and Loop to Face. Then, the Face-to-Face meta-relations are used for the inner $k$ layers. Finally, for the last three layers, the relations are reversed from the original BREP relations: Face to Loop, Loop to Edge, and Edge to Vertex.

\paragraph{Input Features}
We use both the analytic geometry of each BREP topological element and a few computed geometric summary features as the per node input feature vectors for SB-GCN. The analytic features are a one-hot encoding of the parametric geometry function associated with each node, plus a vector of each function's parameters. We exclude parameters for functions without a fixed parameter list size (such as B-Splines and NURBS surfaces) as well as geometry defined in reference to other geometry (like spun surfaces or intersection curves). See Appendix~\ref{ap:datamodel} for a full list parametric functions and their parameters. These more complex topological elements are never referenced for MCFs, so excluding their parameters is not too detrimental. We also exclude the origin parameter because we found that in practice it is frequently far away from the realized geometry once boundaries are taken into account and acted like noise.

To compensate for this lost localization feature, and give useful summarizing information about the more complex surfaces we exclude parameters for, we compute 4 additional geometric summaries. To localize each node in space, we compute its center of mass and an axis-aligned bounding box. We know that the size of each topological entity is a critical feature to determine which fit together, so we explicitly compute surface areas and arc lengths and add those features. Finally, we compute the moment-of-inertia tensor to summarize the geometry of more complex topological entities. We tested adding PointNet derived features to each entity, but found these to be unhelpful given the very low sample count per toplogical entity, and so do not include them.

Prior to extracting analytic and geometric features, we normalize the part geometry by translating both parts to the origins of their respective coordinate systems, and applying a consistent scale factor to both so the largest dimension of either part has unit length.

\begin{figure}[t]
    \centering
    \includegraphics[width=\columnwidth]{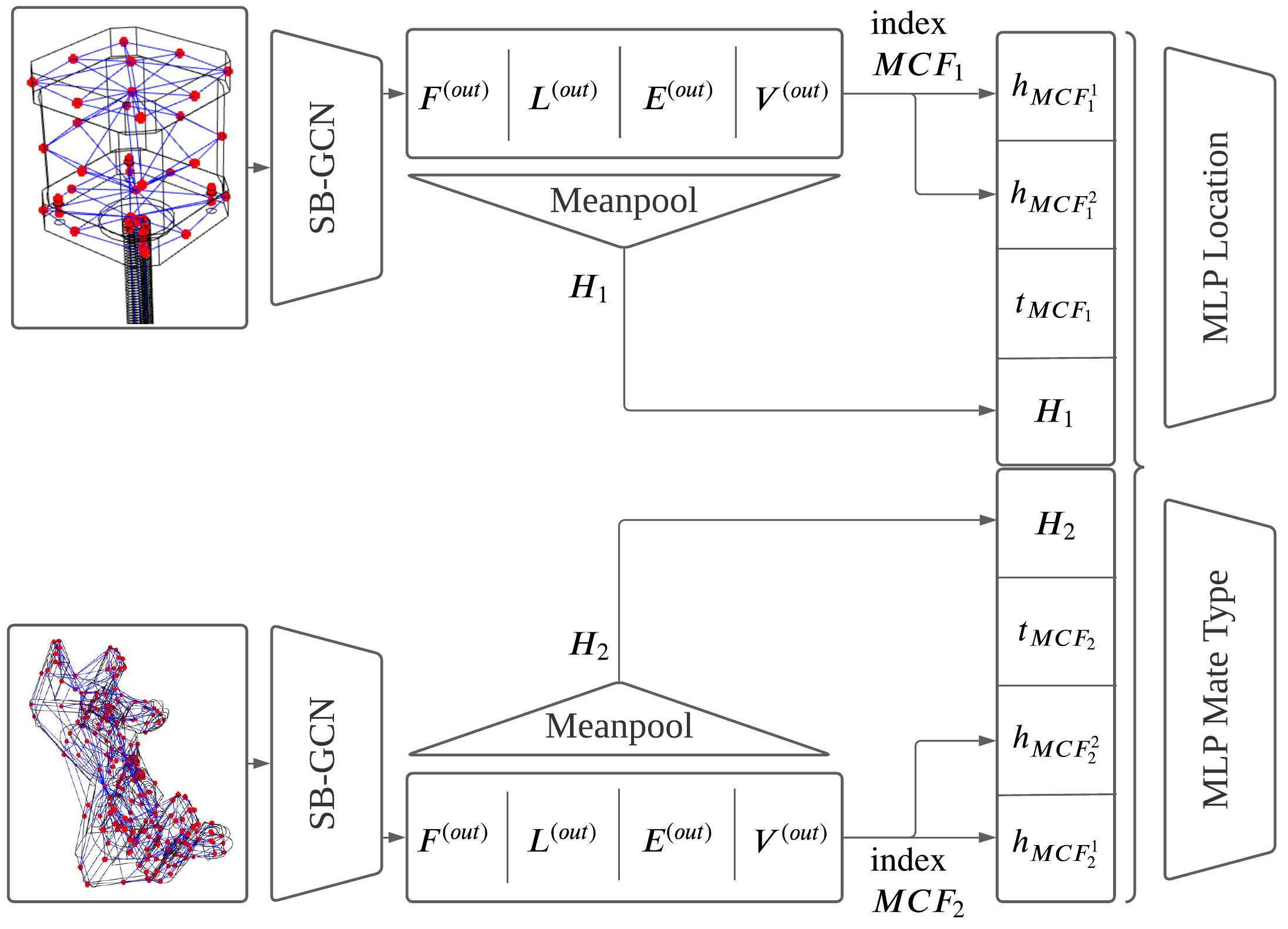}
    \caption{Our classification network. SB-GCN is used to learn embeddings for every topological entity of each part. The embeddings of the topological entities defining each MCF's origin and orientation (written as the topological entities $MCF_i^1$ and $MCF_i^2$ respectively, and thus with embeddings $h_{MCF_i^1}$ and $h_{MCF_i^2}$) are concatenated with a one-hot encoding of the MCF's \emph{origin type}, $t_{MCF_i}$, and a global part feature, $H_i$, computed via a meanpool, to construct MCF features. Each parts' MCF features are concatenated and then classified or scored with an output MLP, depending on the task.}
    \label{fig:classifier}
\end{figure}

\paragraph{Predicting Mates with SB-GCN}
We apply SB-GCN to solve two mating tasks: \emph{location prediction}, which scores and ranks pairs of MCFs adjacent to selected faces on two parts, and \emph{mate type prediction}, which classifies the mate type for a mate with a specific pair of MCFs. We use the same classification network with a different output head for both tasks, shown in Figure~\ref{fig:classifier}. Our classifier uses SB-GCN in a siamese configuration over both input BREPs to learn an embedding for each topological entity. A global part feature is computed for each part with a meanpool of the these embeddings. We then construct a feature vector for each MCF in the mate by concatenating the emeddings of the MCF's orientation and origin topological entities, a one-hot encoding of the origin type, and the global part feature. These MCF features are concatenated and fed to an output MLP for either location prediction (one output) or mate type classification (8 outputs).

\paragraph{Dataset}
Our training data is derived from an earlier version of the dataset described in Section~\ref{sec:dataset}, which has similar statistics, but fewer overall mates. We also applied further filters; we removed mates with overly simple or overly complex geometry (measured by face count) to weed out basic primitives and remove large part outliers that could overwhelm GPU memory.

\begin{figure}
    \centering
    \includegraphics[width=\linewidth]{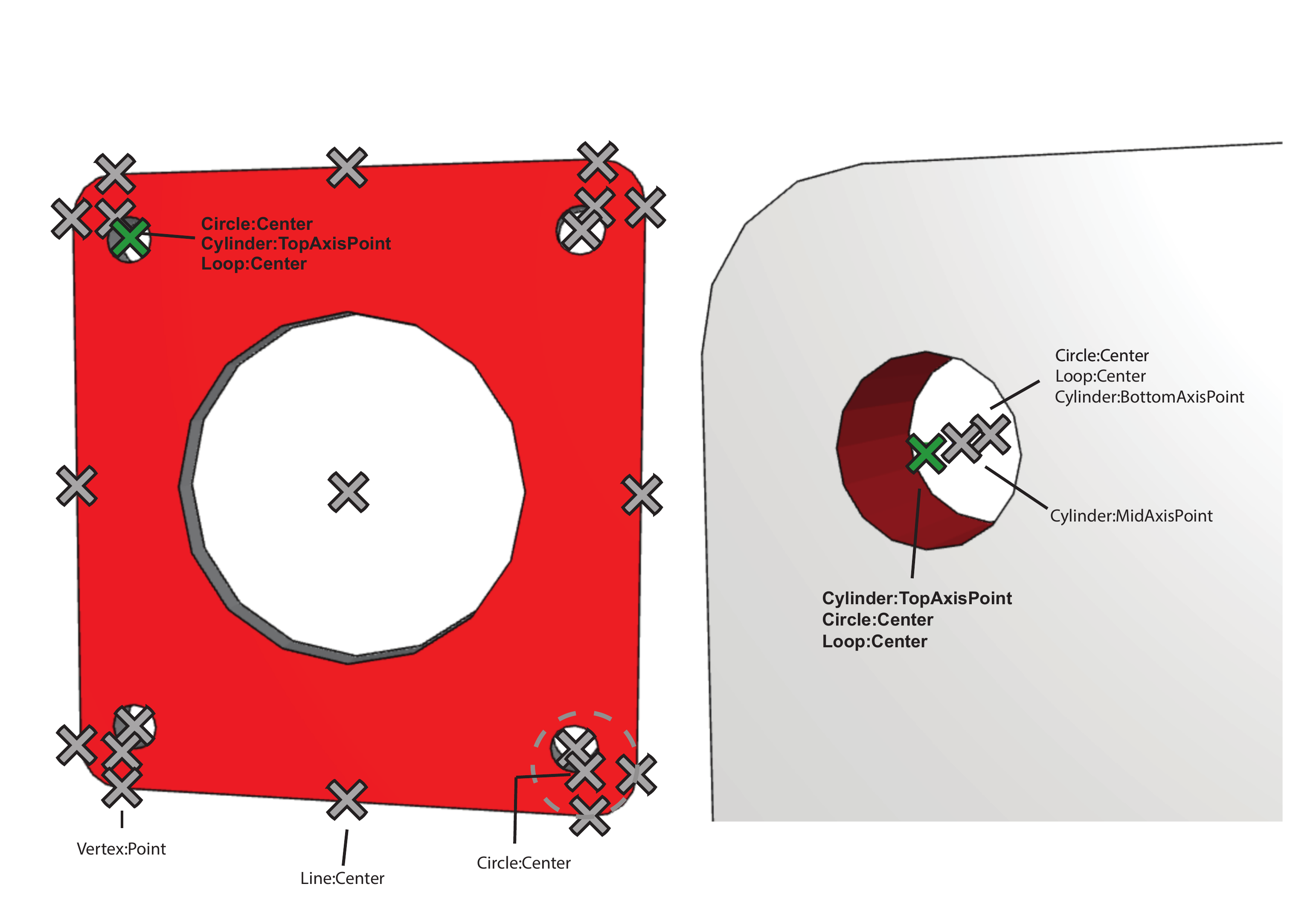}
    \caption{MCF data augmentation. The red face is the selected orientation topology. Green Xs are MCFs equivalent to the ground truth, and grey Xs are other MCFs that can be constructed from those selections, which are used to construct negative location examples. Selected MCFs have been annotated with their location topology and origin types. Note that there are multiple locations aliasing the ground truth MCFs for both selections.}
    \label{fig:dataaug}
\end{figure}

We also augmented our dataset by finding alternative ways to construct equivalent MCFs. Figure~\ref{fig:dataaug} illustrates several ways of constructing the same MCF for a simple part using two different orientation topology selections, as well as other MCFs that can be constructed from those face selections. We augment our dataset by adding mates using all equivalent MCFs as positive examples for mate location, and include MCFs that use the same orientation topology but are not equivalent to construct negative examples. These are grouped by selected topology on each part to create selection examples we train and test on for location. For mate type, we only use the augmented positive examples. Since our UI only supports selecting faces, we remove selection examples that aren't based on a face selection. We apply one final filter to remove MCFs which are not canonically oriented, as described in Section~\ref{sec:overview}, and selection examples with over 10,000 potential mates to limit per-batch memory requirements. After augmentation and filtering we have 267,385 selection examples to train over.

\paragraph{Training}

All models were optimized using the Adam optimizer with a learning rate of 0.001 and betas of (0.9, 0.999). For mate location, we computed an augmented true positive set as described above, labeled as positive examples, and for each compatible potential user selection, all other compatible mates labeled as negative examples. We trained our output classifier with a binary cross-entropy loss, weighted batch-wise. For mate type, we similarly computed an augmented set of true positive mates, labeled with the ground truth mate type (since we are conditioning this model on the mate location), and then trained with cross-entropy loss. Model selection was performed using the average of the hit ratio at $k$: over $k \in [1,10]$ for location and over $k \in [1,8]$ for mate type.

\paragraph{Implementation}
We use Parasolid~\cite{parasolid} to parse BREP files, which we convert to graphs as described above. Our GCN is implemented in Pytorch Geometric~\cite{fey_fast_2019}. We implemented our onshape integration as a browser extension that captures the user's inputs to Onshape, and uses the Onshape API to download BREPs and assembly definitions and to create mates. For SB-GCN we used $n=6$ inner layers, and 64 neurons for every linear layer. 64 neurons were used again for the final predictor MLPs' linear layer.

%% file: sec/6_results.tex
\section{Results}\label{sec:results}

\paragraph{Tasks and Metrics} We evaluated our system's performance in two tasks; predicting the exact mate location conditioned on the user's part and face selections, and predicting the type of a mate conditioned on its ground truth location. Our system is structured as a recommendation system, so as our primary evaluation metric we use hit ratio at $k$;  the likelihood of presenting a correct solution to the user at various depths ($k$) of recommendation list. For the mate location problem in particular, our system as-implemented presents 6 suggestions to the user, so we use this threshold to define our overall system accuracy. We also define the NDCG* as the average inverse log rank of the first correct suggestion to compare methods with a single number. This is a lower-bound approximation to the Normalized Discounted Cumulative Gain commonly used to evaluate retrieval systems, differing in that we don't award additional gain for correct predictions past the first.

\subsection{Baselines}
We evaluate our model against two types of baselines; pointcloud-based deep learning techniques that don't use CAD data, and simple ad-hoc heuristics.

\paragraph{Point Cloud Baselines} We chose to use point clouds as our comparison with discrete geometric representations because they are common in recent assembly modeling work~\cite{sung_complementme_2017,yin_coalesce_2020,huang_generative_2020,li2020learning,vedaldi_pt2pc_2020,yang_dsm-net_2020,mo_structurenet_2019}. We evaluate against three popular methods; PointNet~\cite{qi_pointnet_2017}, Pointnet++~\cite{qi_pointnetpp_2017}, and dynamic graph cnn (DGCNN)~\cite{wang_dynamic_2019}. We structure these predictions following the siamese network design of ComplementMe~\cite{sung_complementme_2017}, using a different output head for each task. Mate type learns an eight class classifier similar to our network, whereas mate location adopts the strategy of existing assembly modeling works and uses regression to predict an offset for each part's MCF; potential mates are then ranked by their MCFs' distance from these predicted offset, effectively snapping to the nearest potential mates. In order to give these algorithms the benefit of the user's face selection, we pre-rotate both parts into their final orientation and center and align the user's selection at the origin for the mate location task; for the mate type task, we transform the pair of parts into their final correct positions, re-centered with the mate at the origin aligned with the first part's mating coordinate frame. We expect that these methods will not do as well as ours because they do not have access to precise analytic geometry of the BREPs, and because they allocate their representation power across the geometry, whereas our method learns representations local to the topological entities on each part that will be interfacing. These models are also much larger than ours; they range from 834k parameters for PointNet to 1.5M for PointNet++, compared to only 126k for our model.

\paragraph{Ad-Hoc Baselines} Our adhoc baselines are task specific. For mate location we have three; \emph{Random}, \emph{Origin Type}, and \emph{Snap to Selection}. \emph{Random} is simply a random ranking of potential mates to get an expected performance floor. \emph{Origin Type} ranks potential mates based on the pair of origin types for their MCFs, in order of the frequency we see such pairs in our dataset. \emph{Snap to Selection} ranks mates by the distance to the center of mass of the user's selected faces; it is intended as a lower bound of the regress-and-snap approach without any regression. For mate type we have one ad-hoc baseline: \emph{Label Distribution}. This always ranks mate type according to their frequency in our test set, ignoring the input.

\subsection{Data Ambiguity}

\begin{figure*}[t!]
    %\centering
    \includegraphics[width = .9\textwidth]{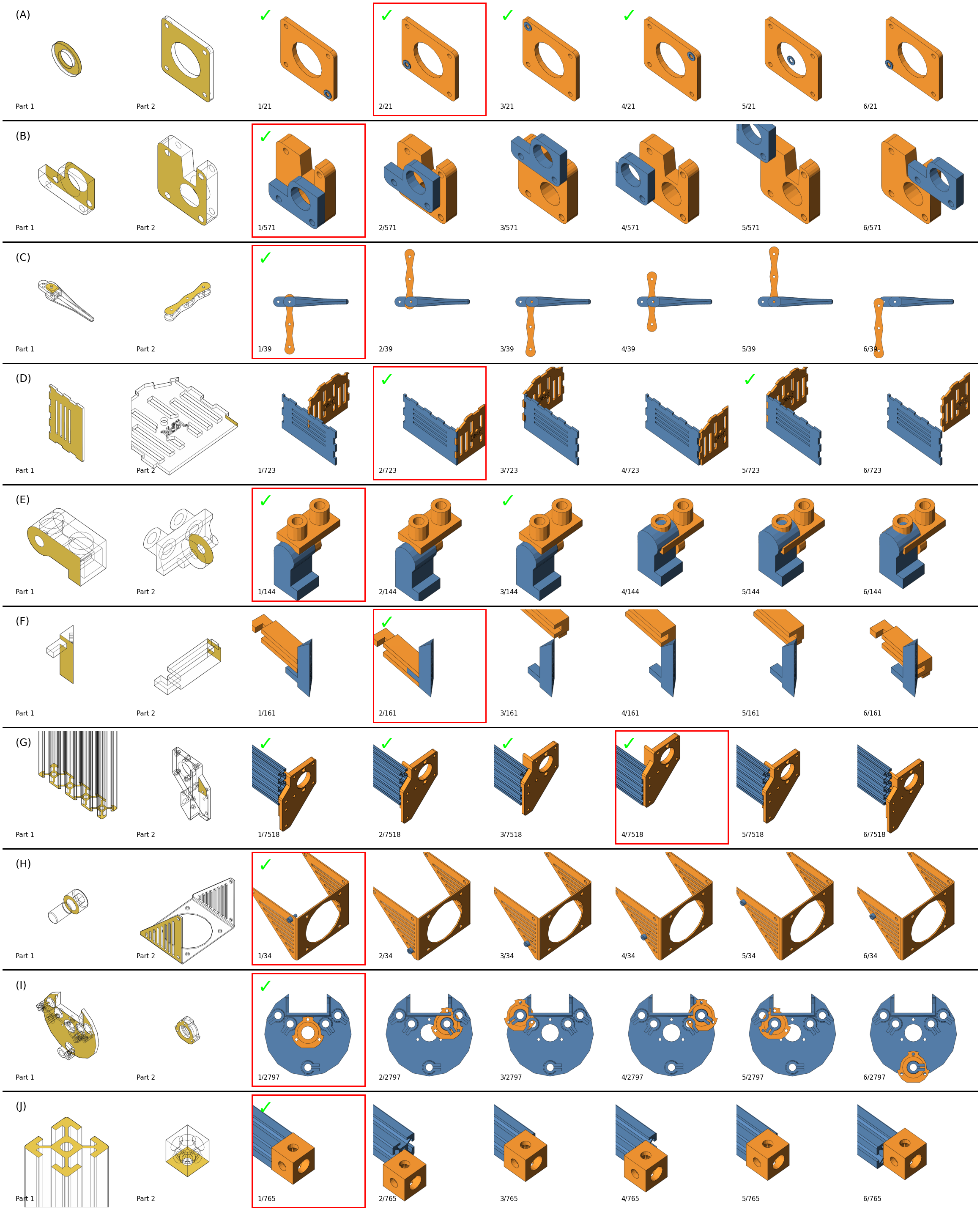}
    \caption{Location predictions. Mates labeled as correct in our dataset are boxed, and mates that a CAD expert judged as good prediction have a checkmark. Prediction rank is indicated below. For these examples, we have excluded suggested mate locations that are equivalent to an earlier mate given the mate type to better reflect user preferences, which affects both rank and number of possibilities compared to our quantitative results.
    }
    \label{fig:locationgallery}
\end{figure*}

\paragraph{Location Ambiguities} 
Mate placement is a subjective task that depend on the user's design intent and the context they are working within. Our dataset only contains the mates that were needed to create the objects that the parts were used for, but parts can be used in other ways not seen in our dataset. In our quantitative metrics we evaluate strictly against the mates which appear in our dataset, so we asked a CAD expert to evaluate some of our system's prediction and indicate which suggestions were plausible uses of those parts. For all examples shown in the paper (Figures \ref{fig:locationgallery} and \ref{fig:locationfailures}) we have indicated with a green checkmark all mates that the CAD expert selected.

\begin{figure}
    \centering
    \includegraphics[width = 0.5\linewidth]{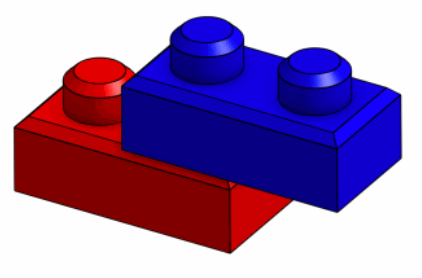}
    \caption{Example of an ambiguous mate. The survey participants were divided between cylindrical (blocks can be rotated around the peg and slid off the peg), slider (blocks can be slid off the peg), and fastened (blocks should stay in place). In the dataset, this is classified as a parallel mate. }
    \label{fig:ambiguousmate}
\end{figure}

\begin{figure}[t]
    \centering
    \includegraphics[width=\linewidth]{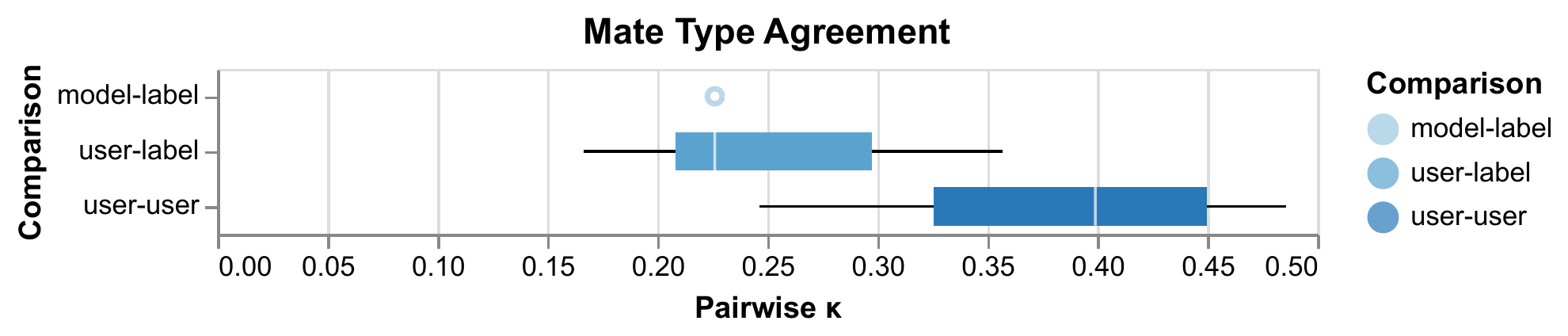}
    \caption{
    Mate type agreement, measured by pairwise Cohen's $\kappa$, between surveyed CAD users, our model, and the dataset labels. Our model agrees with the labels to the same degree as the median expert, indicating that we achieve human performance. The experts agree within themselves to a greater degree than they agree with the dataset labels, indicating that there are additional contextual clues for mate type not captured by just looking at a pair of parts.
    }
    \label{fig:agreement}
\end{figure}

\paragraph{Mate Type Ambiguities}
The mate type that a design uses is highly dependent on the designer's context. For example, our dataset over-represents \emph{fastened} mates, but visual inspection finds gears marked as fastened rather than revolute; the user who created that assembly was not concerned with modeling the gear motion. To quantify the level of ambiguity on our dataset, we recruited seven expert CAD users to label 96 mates (12 of each mate type as labeled in our dataset) with what mate type they thought it should be. Overall, there was a majority consensus among the CAD users for 70 out of the 96 mates, which suggests we can expect to an upper bound of roughly 73\% for top suggestion accuracy. Figure~\ref{fig:ambiguousmate} shows an example of a mate that our experts gave three different labels to, and the dataset gave a fourth. We further quantified agreement using pairwise Cohen's $\kappa$, a 0-1 measure of agreement above random chance pictured in Figure~\ref{fig:agreement}. We measured agreement among the CAD experts, between the CAD experts and our dataset, and between our dataset and our model predictions. This analysis showed that while our CAD experts agree more with each other than with the dataset labels, our model's predictions agree with the dataset to the same degree as the median CAD expert, indicating that it achieves human-level performance.

\subsection{Mate Location Prediction}

 Figure~\ref{fig:locationgallery} shows a selection of location predictions from our system across a variety of complexities from a dozen possibilities to several thousand. Our model has learned to align key features like holes and edges of similar sizes. In example (H) we see the effect of not knowing the mate type in advance; the bolt holes and slots have a similar radius and our model suggests mates that could be either \emph{revolute} (in the bolt holes) or \emph{pin slots} (in the slots). Examples (E) and (I) show that our model learns to align features but does not learn to avoid part overlap.

\begin{figure}[t]
    \centering
    \includegraphics[width=\linewidth]{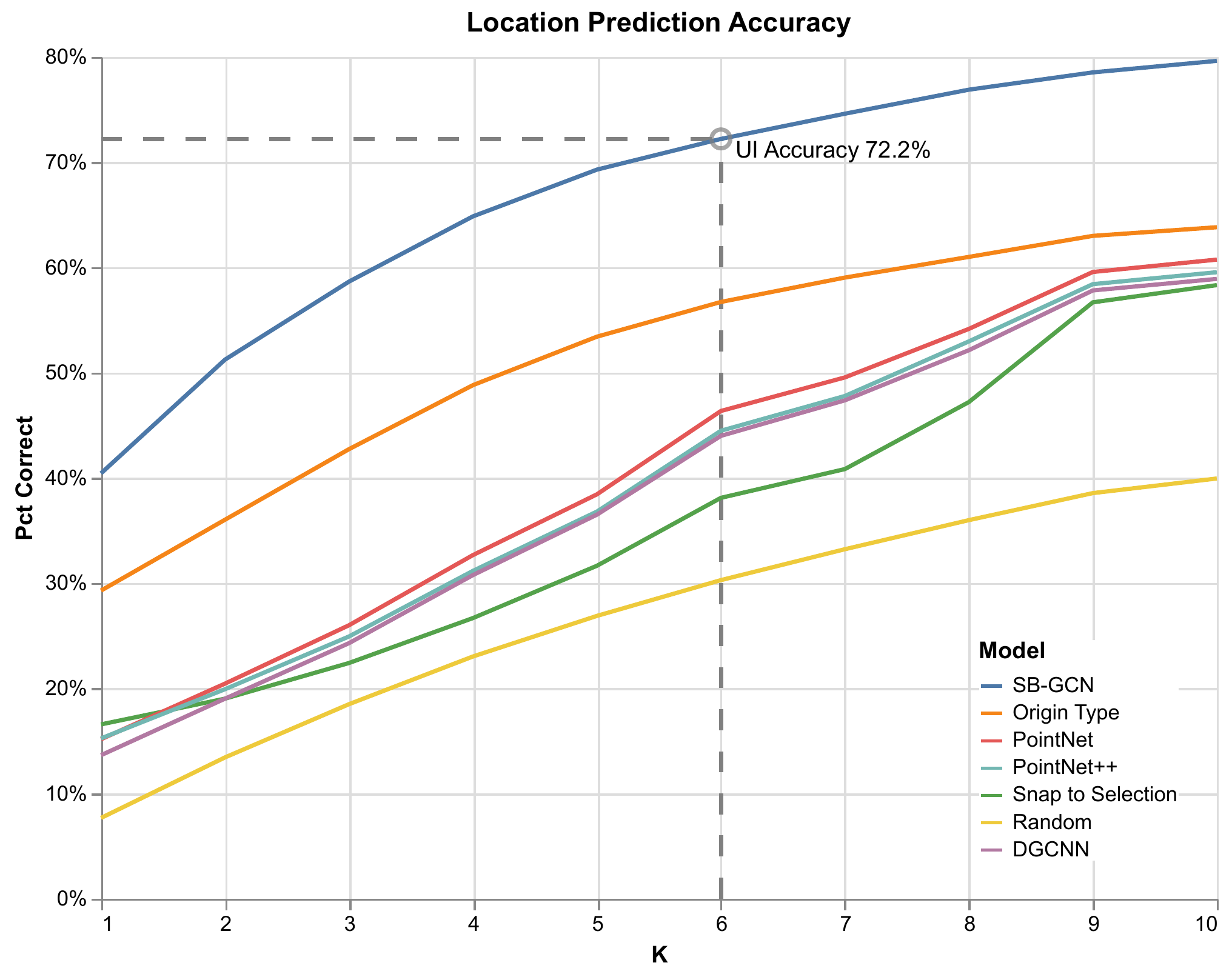}
    \caption{Our model compared against baselines on mate location prediction task.}
    \label{fig:locationresult}
\end{figure}

\paragraph{Baseline Comparisons}
Figure~\ref{fig:locationresult} illustrates the performance of our model in predicting mate location, as compared to several baselines. Our model outperforms all baselines, and achieves a 72.2\% accuracy at our UI threshold of 6 predictions. The inference type baseline, our simplest model using CAD data, lags behind our system by roughly 15\%, but still in turn outperforms all geometric baselines. While these geometric baselines do learn something; they perform better than a random selection baseline, they do not significantly outperform Snap to Selection, a model with no learning.

\begin{figure}[t]
\centering
\includegraphics[width = .5\textwidth]{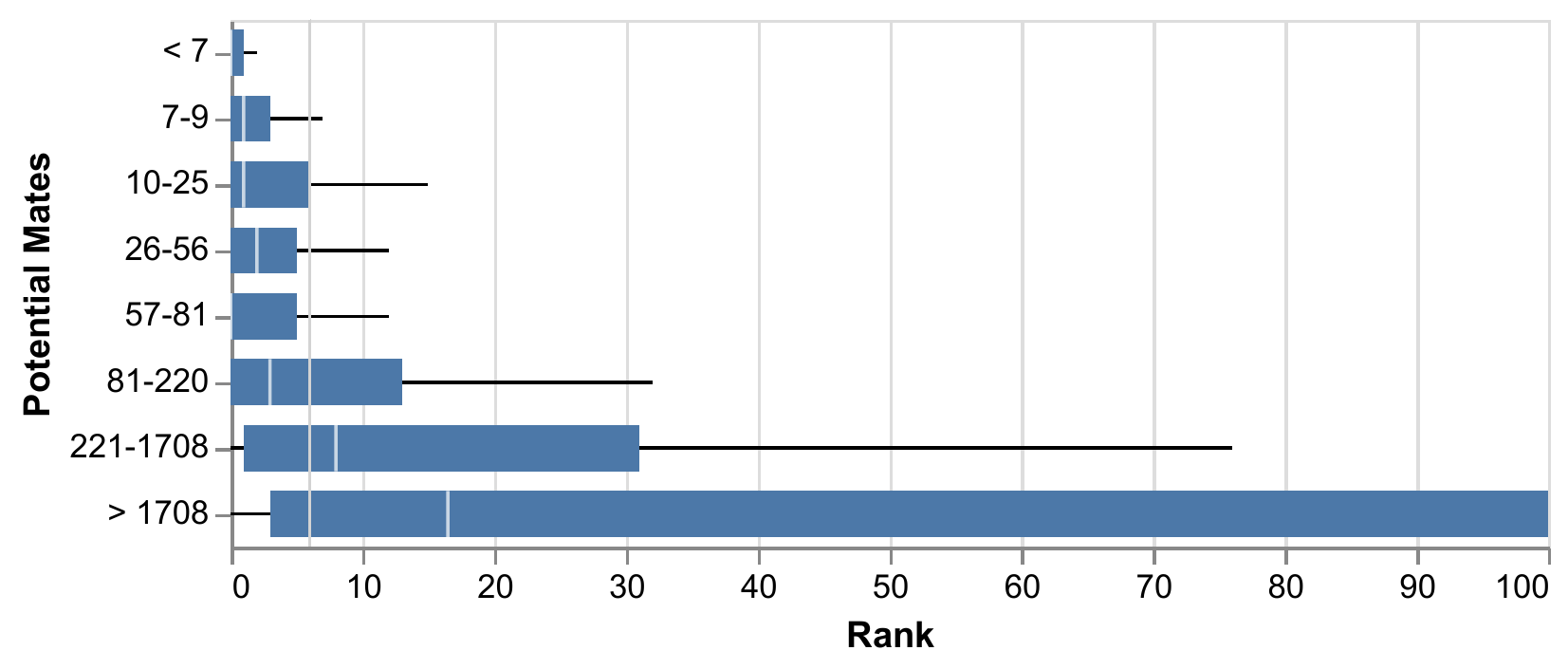}
\caption{Rank of the first correct mate suggestion as the number of potential mates grows. Ranges selected to have roughly even mate counts (~3500), except the final bin which has only 646. While variance does grow with the number of potential mates, the median correct prediction rank grows much more slowly, and only exceeds our UI cutoff of 6 selections (light grey line) above 220 possibilities. Outliers have been excluded from whiskers, and the whisker for the final bar has been truncated.}
\label{fig:complexity}
\end{figure}

\begin{figure*}[t]
    %\centering
    \includegraphics[width = .925\textwidth]{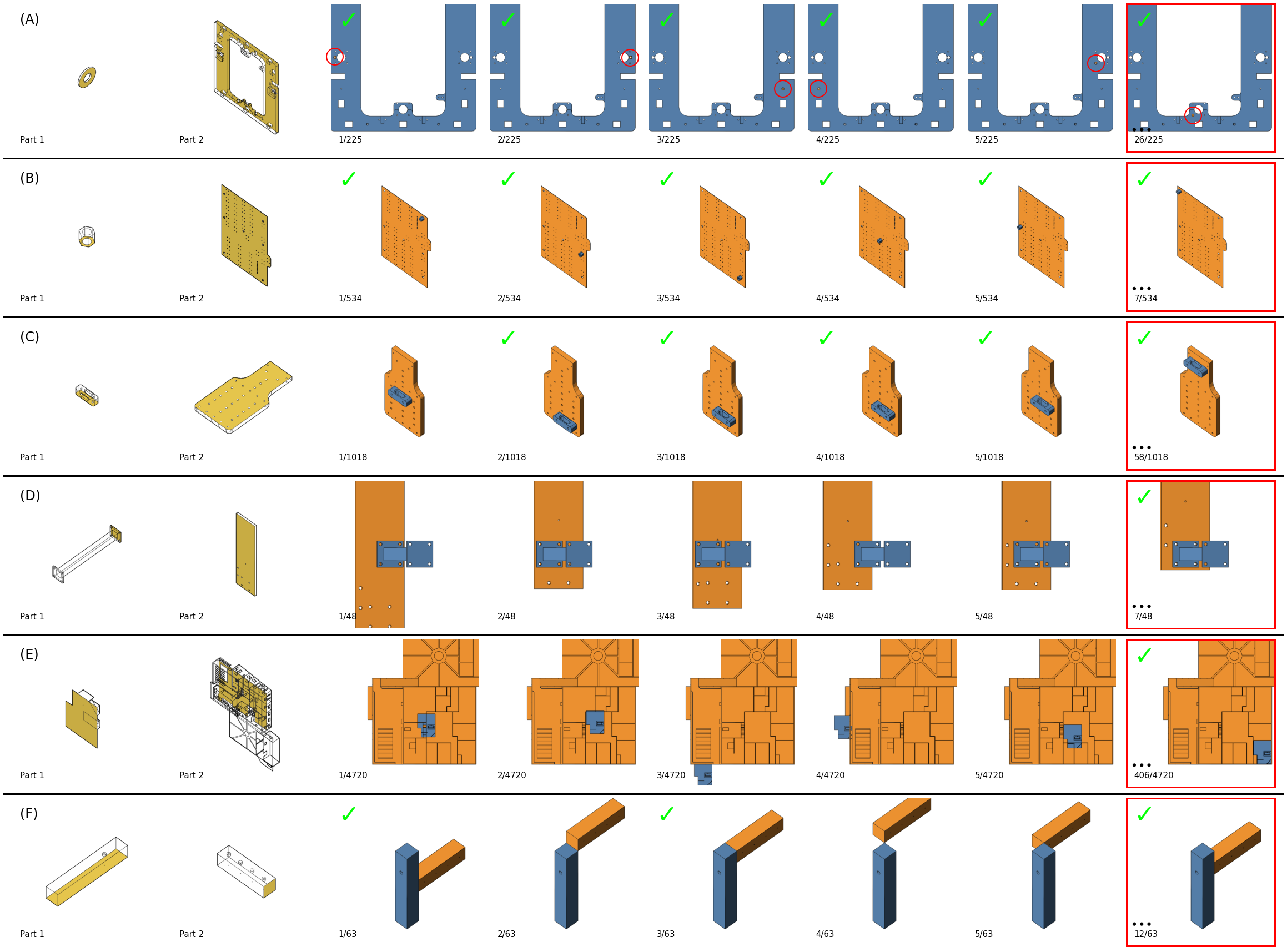}
    \caption{Various failure cases. (A)-(C) highlight cases where our model produces plausible results, but does not have the greater context to understand the user's intent. (D) and (E) are cases where a better understanding of extrinsic geometry would help: in (D) to understand multiple-hole alignment, and in (E) to capture geometrically close features (the walls), which are topologically distant from the pertinent MFC reference (the floor). (F) shows a case where the material context would help; the are square lumber, and the original creator intended to align one piece with a pre-drilled nail hole in the other.
    }
    \label{fig:locationfailures}
\end{figure*}

\paragraph{Scaling}
We evaluate how well our method scales with the number of potential mates to rank. We binned prediction results by the number of potential mates (without the merging present in Figures \ref{fig:locationgallery} and \ref{fig:locationfailures}, so these numbers are higher than appears in those figures), then computed quantile statistics of the rank of the first correct prediction, which we plot in Figure~\ref{fig:complexity}. The median rank grows much less quickly than the number of possible mates, and in fact stays below our interface cutoff of six suggestions for six of the eight buckets, showing that our predictor can scale well to more complex problems.

\paragraph{Failure Cases}
Figure~\ref{fig:locationfailures} illustrates some failures of our model. The far right column shows the first correct prediction and its rank. Examples (A)-(C) illustrate cases where additional context beyond the parts would be needed to correctly infer the mate location. In each of these cases, a third part interacts with the mated part. Our model did, however, pick alternatives that the expert CAD user validated as useful for some use case -- just not the one seen in our dataset. Example (D) shows a common problem when parts have multiple locations that need to be aligned, our model does not always align all of them; our chosen locations align one, two, or three holes, but miss the one location that would align all four. In example (E), the walls as modeled are actually hovering above the floor, and so are not topologically close enough to be captured in MCF representations. Both of these examples suggest we should incorporate more extrinsic geometric features and connectivity to account for mismatch between topological and geometric distances. In the final example, the horizontal board has no marks on its end, so our model attempts to match it with the midpoint and a similarly sized edge of the vertical one. The correct mate aligns the end of the horizontal board with a pre-drilled nail hole in the vertical board, and since it's a nail, there is no hole modeled in the endcap to suggest alignment.

\subsection{Mate Type Prediction}

\begin{figure}[t]
    \centering
    \includegraphics[width=\linewidth]{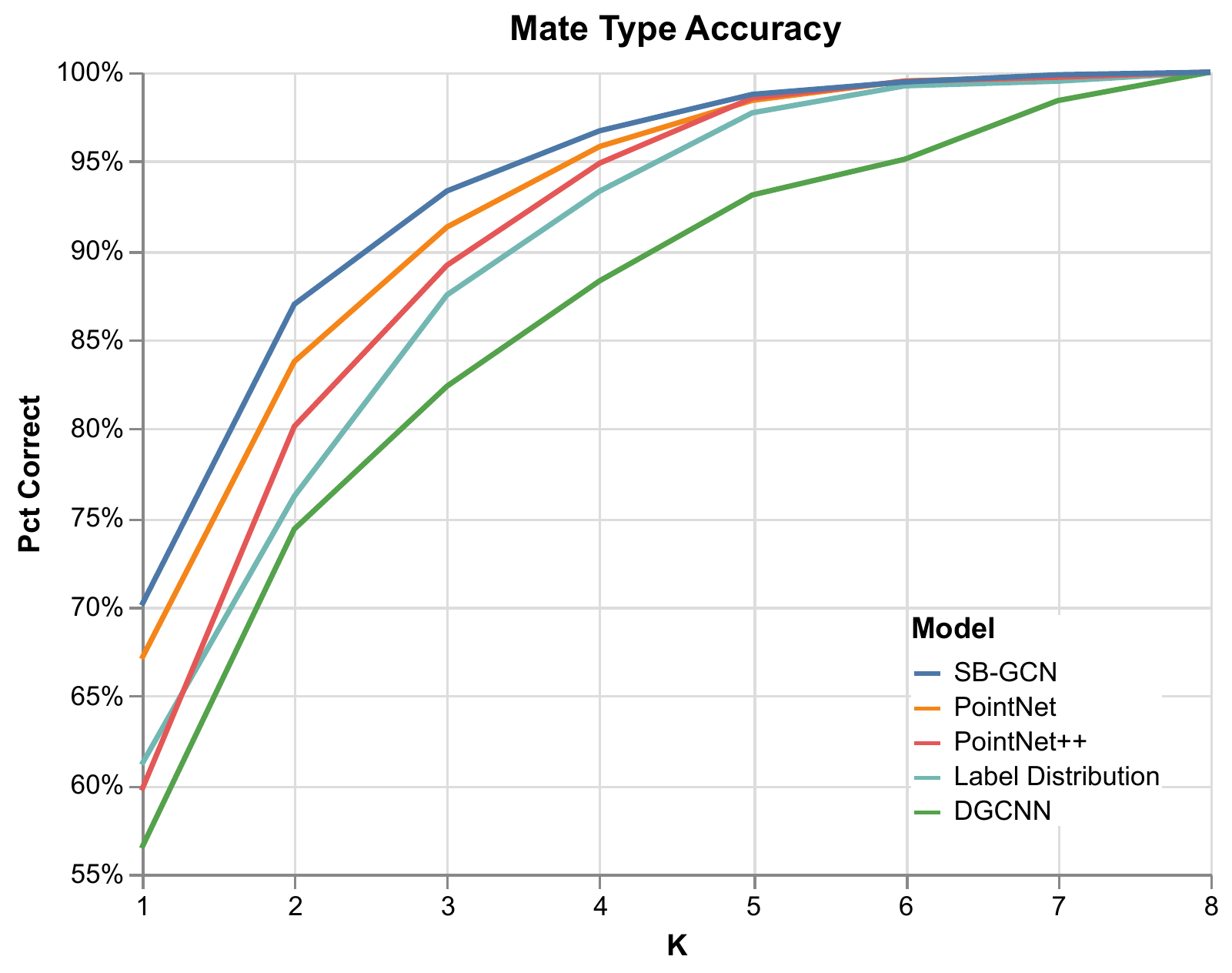}
    \caption{Our model compared against baselines on mate type prediction task.}
    \label{fig:typeresult}
\end{figure}

Figure~\ref{fig:typeresult} illustrates performance against baselines for the mate type task. Our model outperforms all of the baselines, though the gap is not as significant as in the location task due to the tighter bounds between the label distribution baseline (presumed lower bound) and our ambiguity-based estimate of 73\% first suggestion accuracy upper bound. Having access to definitive analytic information, for example that the MCFs are centered on a cylinder and a circle indicating likely rotational degrees of freedom, gives our model an edge over the pointcloud based methods that must infer this from samples. The 73\% upper bound estimate is very rough due to a small sample size, so our model's 70\% top suggestion accuracy and alignment with the median human expert (discussed above, see Figure~\ref{fig:agreement}) leads us to believe our model is approaching the limits of accuracy given the dataset ambiguity.

\subsection{Ablations and Design Decisions}\label{sec:ablations}
\begin{figure}[t]
\centering
\includegraphics[width=.5\textwidth]{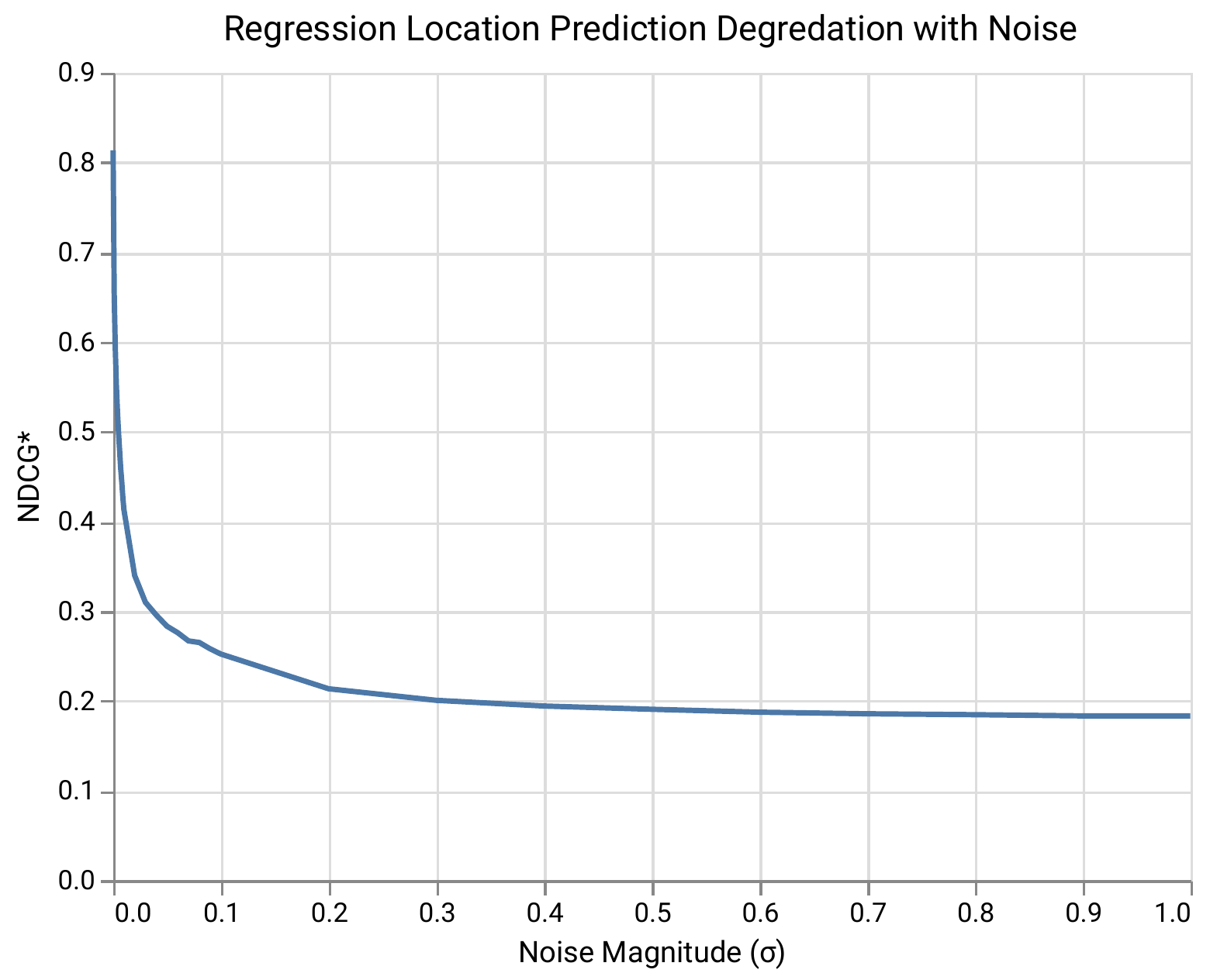}
\caption{Degradation of regression-based location accuracy in the presence of noise. The sharp accuracy drop-off in the presence of small prediction noise shows that regress-and-snap prediction cannot perform well for predicting mate coordinate frames.}
\label{fig:noise}
\end{figure}

\paragraph{Analytic versus Discrete Geometry}
Since we want to integrate with existing BREP based CAD software, and need the precision of analytic geometry to interface with CAM systems, any mate definitions we produce will need to be defined in reference to BREPs. However, there is still a question of whether to use the BREP data directly for prediction, or to convert to a discrete geometric representation, like a point cloud or mesh, that is more commonly used in assembly modeling. In this alternative, we would predict relative offsets for each part, then project to the nearest BREP mate. Our baseline comparisons show that this approach does not work well for pointcloud based models that have previously been used in assembly modeling. To test the viability of the regress-and-snap strategy in general, we tried removing the learning component by testing against a noisy oracle that always predicts the true offsets, plus a controllable noise vector scaled to be a fraction of the standard deviation of offsets off all mates under consideration. Figure~\ref{fig:noise} shows that regress-and-snap is a losing strategy: accuracy drops at even very small amounts of output noise.

\paragraph{Feature and Network Selection}
To determine which features to give our network, we incrementally added layers of features; just parametric functions, parametric function parameters, topological entity size, and bounding box plus center of mass and moment of inertia tensor. We also experimented with several convolutional architectures. The simplest was our \emph{plain} network, which does not use the structural BREP data at all, instead learning embedding features with a shared MLP across topological entities. Our most direct analog to \cite{willis_fusion_2020} is our \emph{homogeneous} network. Similar to Willis et. al., we treat the BREP as an undirected graph and apply homogeneous message passing. However, since we need embeddings of all entity types, not just faces, we include all classes of entities in our graph. We also could not use grid-sampled face points as in Willis et. al. since non-face entities cannot be grid sampled. We tried adapting other sampling strategies to topological entities of lower dimensionality, but this strictly decreased performance and so was dropped. Since our BREP graphs are naturally heterogeneous, we also tried a heterogeneous variant of this network that learned a different message passing function for each type of edge (vertex-edge, edge-loop, loop-face, face-loop, loop-edge, and edge-vertex). This approach is more costly as it has six times the number of parameters. Finally, we tried the BREP structured network described in Section~\ref{sec:methods}, which learns separate message passing functions for each edge type while having the same number of parameters as the homogeneous network, and fewer overall message passes than either homogeneous or heterogeneous. We found that these two axes of exploration, features and network convolution, both improve results for the location task independently, and that each can compensate for the other. Table~\ref{tab:simplifiedablation} summarizes these differences. We hypothesize that the additional parameter and computed features capture some of the information that the network otherwise needs neighborhood data to infer. We choose to use our structured model with all features since it performs as well as other architectures that are more costly in computation and parameter count.

\begin{table}[t]
    \centering
    \caption{Validation NDCG* for mate type and location problem for experiments with or without a GCN, and with all input features or just the parametric function type.}
    \begin{tabular}{c|c c | c c|}
         & \multicolumn{2}{c|}{Mate Type} & \multicolumn{2}{c|}{Location} \\
         & Fn Type & All  & Fn Type & All  \\
         \hline
         No GCN & .832 & .846 & .470 & .561\\
         SB-GCN & .846 & .850 & .547 & .576
    \end{tabular}
    \label{tab:simplifiedablation}
\end{table}

%% file: sec/7_discussion.tex
\section{Limitations and Future Work}

In proposing the first learning model for suggesting CAD mates, our approach leaves several avenues of future work.

\begin{figure}[h]
    \centering
    \includegraphics[width=0.4\textwidth]{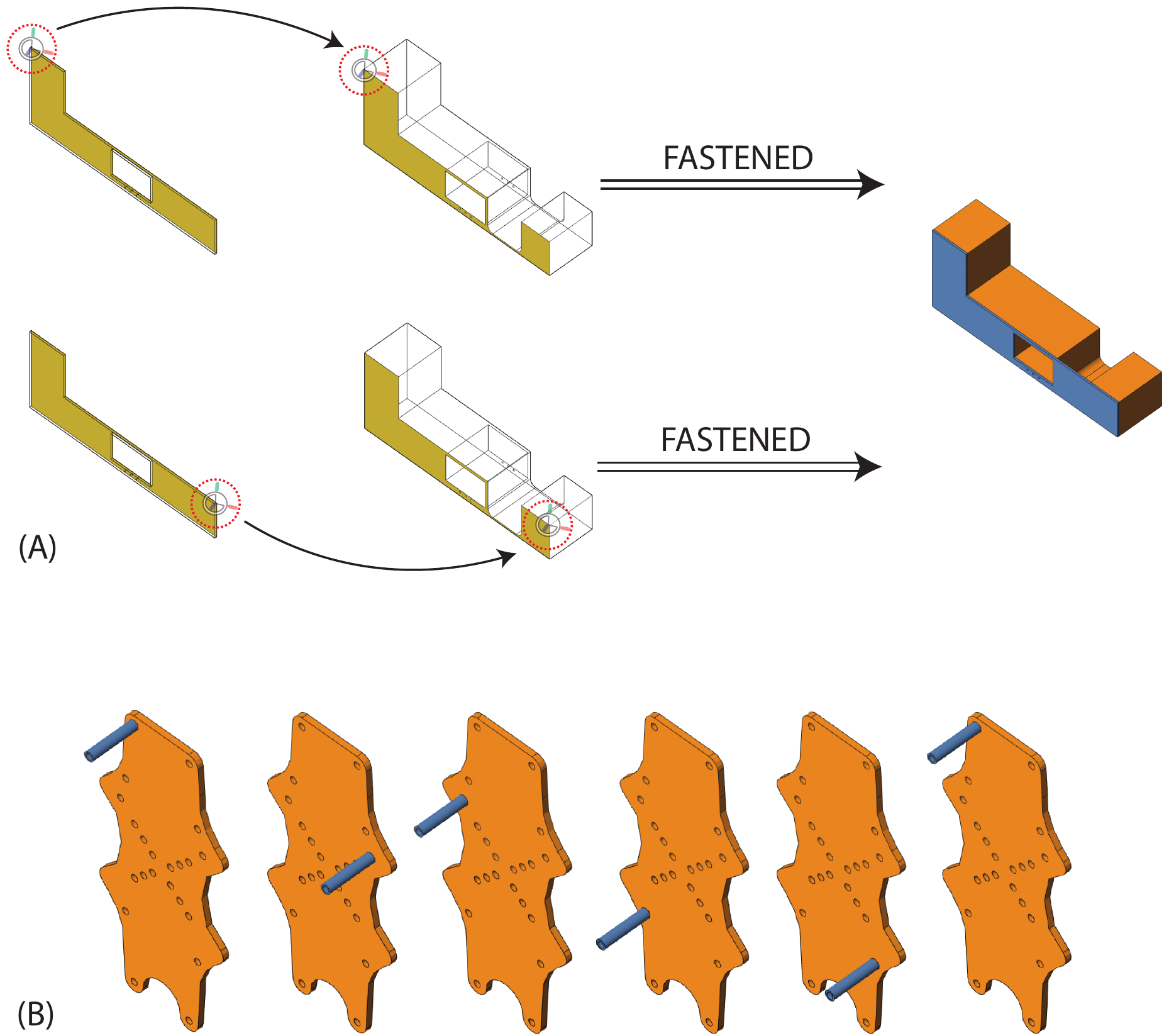}
    \caption{Two types of false negatives present in our training data. (a) The same relative degrees of freedom far a part pair can be achieved by different combinations of MCFs, dependent on the mate type; we currently only capture those with equivalent MCFs. (b) Identical parts pairs appear in multiple configurations in our dataset, but since we train and validate our model \emph{per-mate}, these alternatives appear as false negatives. If we can match topological entities across equivalent BREPs, we could unify these examples.}
    \label{fig:falsenegatives}
\end{figure}

 The inevitable existence of false negatives in our dataset means that our quantitative results under-represent the quality of our predictions, as evidenced by the expert annotations in Figure~\ref{fig:locationfailures}. While no \emph{in the wild} dataset can ever capture all ways that a user may wish to use a given part pair, data augmentation can be used to reduce false negatives. In this work we augmented our dataset at the MCF level by computing alternative constructions for the same coordinate frames. Future work could extend this in three ways. First, as shown in Figure~\ref{fig:falsenegatives} (a), merging mates with different MCF locations that combine to the same relative positions if the mate type is given gives us a more accurate picture of result quality (as done in Figures~\ref{fig:locationgallery} and \ref{fig:locationfailures}). While this cannot be done as a post process if mate type is not known at inference (as reported in Figure~\ref{fig:locationresult}), additional work could be done to augment the database finding all pairs with all equivalent MCFs. Second, as shown in Figure~\ref{fig:falsenegatives} (b), our work does not capture the alternative uses of pairs of parts that exist in our database because we separate our training and validation examples by mate pair rather than part pair. Unifying these and separating our examples by part pair would require extending our fingerprint for identifying equivalent BREPs to a full BREP matching including a mapping between BREP topological entities. Finally, there are symmetries that we could exploit to augment our data with additional plausible mates; in Figure~\ref{fig:dataaug} this would result in the center MCFs of each hole being marked as ground truth since that part has 4-fold rotational symmetry. 

Several additional features are required to make our system viable in a CAD workflow. Additional user interactions to select non-face topologies would remove the restriction of only selecting face orientations. Some mate suggestions are clearly non-viable due to intersections or displacements, these could be heuristically pruned to improve suggestion quality. We would also train an additional categorical predictor for mate location to allow for non-canonical part orientations, which together with non-face selection would allow our system to work with the vast majority of mates.

We would also like to include some discrete geometry features into our model. While point density around sharp features presents a challenge for PointNet features, other models such as MeshCNN \cite{hanocka_meshcnn_2019} or the curve and surface convolution features of UV-Net \cite{jayaraman2021uv}. Further, incorporating the greater assembly context into our predictions is an exciting challenge  that we leave to future work (see Figure~\ref{fig:locationfailures} (A)-(C)).

Our system is built to work with the Parasolid modeling kernel and with MCF based mates, but CAD systems use a variety of modeling kernels and mating schemes. Adapting our approach to other modeling kernels is straightforward; the primitives for which we support parametric features and MCF references are common to every BREP modeling kernel, and other entity types are only featurized by a categorical variable and geometric summary information. Adapting our data set to other modelers may incur data loss since different primitives may be used to generate the same geometry, leading to different potential MCFs. We will initially release our dataset and code in both Parasolid format and a pre-featurized serialization to avoid dependence on a modeling kernel, and plan to also make it available in STEP format with an OpenCascade implementation of BREP featurization.

Finally, this work lays the machine learning foundation for automatic mating interfaces and leaves open exciting avenues for expanding the capabilities of these systems. We limited our mate location search in this work to the neighborhood of the user's selection both to incorporate the user's design intent, and to limit the number of mates under consideration. If we expand the  number of MCFs under consideration in each part, the number of possibilities grows quadratically, and will exhaust available memory at training time. This could be somewhat alleviated by random negative sampling, but even in the one-face neighborhood we had to cap each example to at most 10k MCF pairs to limit memory consumption, and some face selections can result in millions of pairs. Future work can address this limitation by making the dependency linear, choosing a constant number of MCFs on one selected part as the mate location intent, then considering all MCFs of the other part. Another direction is to  extend the system to include part selection---given an MCF of a single part, suggest other parts that could mate at that location. A system combining part retrieval with placement has the potential to drastically speed up the CAD mating workflow.

\section{Conclusion}
In this work we proposed SB-GCN the first heterogeneous graph network for BREPs capabable of learning embeddings for many types of topological entities. We used SB-GCN to build the first assembly modeling tool aligned to the unique needs of the CAD workflow; it takes BREPs as input, is category agnostic, models assemblies as a collection of constraints, and defines those constraints in reference to analytic geometry. While we applied our network to problems of mating, and to one particular model of mates, we designed our model to learn to represent the core building block of all BREP-based CAD software: topological entities. Because of this, SB-GCN should be applicable to other mating systems, and other CAD tasks. Finally, we collected and cleaned the first BREP-based assembly modeling dataset, which we are releasing in conjunction with our model and baseline tasks. We are excited to see what new innovations in CAD will be built on this foundation.

%% file: sec/8_appendix.tex
\pagebreak
\appendix
\section{Data Model}\label{ap:datamodel}

\paragraph{Node Types}
Table~\ref{tab:parametrictypes} summarizes the parametric functions in our system, along with their parameters. We exclude the origin parameter because these do not need to correspond to the topological entity location once boundaries are taken into account, and in practice are often far from surfaces, resulting in noise.

\begin{table*}[h]
    \centering
    \caption{Parametric functions and their parameters.}
    \begin{tabular}{c|c|l|l}
         Function & Topology Type & Parameters & Description  \\
         \hline
         plane & face & \st{origin}, normal & A plane.\\
         cylinder & face & \st{origin}, axis, radius &  A cylinder.\\
         cone & face & \st{origin}, axis, half-angle & A cone.\\
         torus & face & \st{origin}, axis, major-radius, minor-radius & A torus.\\
         bsurf & face & & A NURBS surface.\\
         swept & face & & Surface swept along a curve.\\
         spun & face & & Surface of rotation about an axis.\\
         blend & face & & Surface blending two other surfaces.\\
         outer loop & loop & & The outer trimming curve of a face.\\
         inner loop & loop & & An inner trimming curve of a face.\\
         line & edge & \st{origin}, direction & A line.\\
         circle & edge & \st{origin}, normal, radius & A circle.\\
         ellipse & edge & \st{origin}, normal, major axis direction, major axis, minor axis. & An ellipse.\\
         bcurve & edge & & A B-spline curve.\\
         icurve & edge & & The intersection of two surfaces.\\
         spcurve & edge & & Curve embedding in a surface.\\
         tcurve & edge & & Trimmed curve.\\
         pline & edge & & A polyline.\\
         point & vertex & \st{origin} & A point.
    \end{tabular}
    \label{tab:parametrictypes}
\end{table*}

\paragraph{Origin Types}
Each MFC has an \emph{origin type} that specifies how the origin is computed from its reference topological entity.  Table~\ref{tab:origintypes} lists all of the available origin types, along with the types of geometry they are applicable. Geometry without supported origin types are not able to be referenced to define MFCs. Note that all of these computations except \emph{center} and \emph{point} rely on boundary information, and so are not computable from the parametric geometry a topological entity in isolation, highlighting the need to integrate boundary information to properly understand the mating coordinate frames. Also note that the origin type is highly correlated with the geometry type of the referenced topological entity, which is why it is a useful proxy for our Origin Type baseline.

\begin{table*}[h]
    \centering
    \caption{Supported origin types for mating coordinate frames.}
    \begin{tabular}{r|l|l}
         Origin Type & Supported Functions & Description  \\
         \hline
         center & circle, ellipse, loop & The center of a curve or inner loop.\\
         centroid & plane & The centroid of a planar face.\\
         mid point & line & The mid point of a line segment.\\
         point & point, cone & The location of a point or vertex of a cone.\\
         top axis point & cylinder, torus, cone, spun & The most positive point of a surface of revolution projected onto its axis.\\
         mid axis point & cylinder, torus, cone, spun & The mid-point of a surface of revolution projected onto its axis.\\
         bottom axis point & cylinder, torus, cone, spun & The most negative point of a surface of revolution projected onto its axis.
    \end{tabular}
    \label{tab:origintypes}
\end{table*}

\paragraph{Computing MCF Orientation}
The orientation of an MCF is defined by reference to a supported topological entity. Table~\ref{tab:orientationcomp} lists the geometric functions from which orientations can be determined, and describes how they are computed.

\begin{table*}[b]
    \centering
    \caption{Z-axis computation for supported topological entities.}
    \begin{tabular}{c|c}
         Topological Entity Function(s) & z-Axis \\
         \hline
         plane &  normal\\
         cylinder, cone, torus, spun & axis parameter\\
         inner loop & plane normal if inner loop of a plane\\
         line & along edge in positive parameterization direction\\
         circle, ellipse & plane normal if adjacent to a plane, otherwise axis parameter\\
         point & z-axis of CS part was modeled in.\\
    \end{tabular}
    \label{tab:orientationcomp}
\end{table*}

The orientation reference of an MCF only determines the frame's z-axis. The y-axis is computed as the cross product of this z-axis with the x-axis of the coordinate from that the part was modeled in, or the cross product with it's y-axis if that x-axis is parallel to the chosen z direction.

\section{Ablations}
\begin{figure*}[t]
    \centering
    \includegraphics[width=\textwidth]{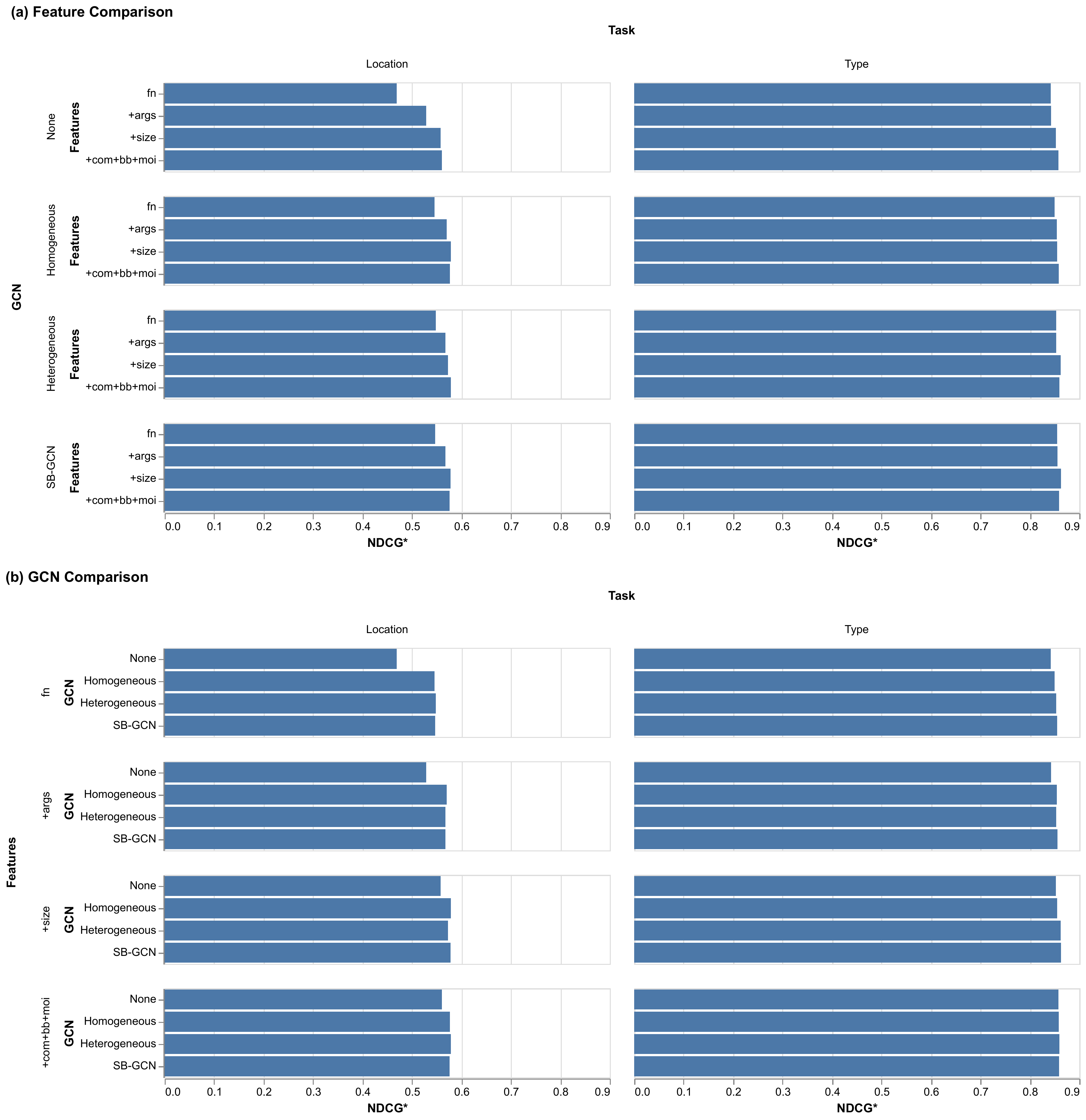}
    \caption{Validation NDCG* for GCN type and feature set experiments trained against 70k subset. NDCG* is a modified NDCG score that stops counting at the first correct mate. For mate type this is equivalent to NDCG since there is only one correct mate type.}
    \label{fig:ablations}
\end{figure*}

We tried sixteen combinations of graph architecture and feature set across the mate location and mate type problem. Figure~\ref{fig:ablations} shows all of these results, compared along either axis. The mate type problem does not show much difference between methods, but for mate location there is a clear trend that having a GCN is better than not, that more features are better, and that adding features can partly make up for not using a GCN.

